\documentclass{article}

 \usepackage[eandd, final]{neurips_2026}
\usepackage[utf8]{inputenc} % allow utf-8 input
\usepackage[T1]{fontenc}    % use 8-bit T1 fonts
\usepackage{hyperref}       % hyperlinks
\usepackage{url}            % simple URL typesetting
\usepackage{booktabs}       % professional-quality tables
\usepackage{amsfonts}       % blackboard math symbols
\usepackage{nicefrac}       % compact symbols for 1/2, etc.
\usepackage{microtype}      % microtypography
\usepackage{xcolor}         % colors
\usepackage{xspace}
\usepackage{hyperref}
\usepackage{etoolbox}
\usepackage[most]{tcolorbox} % 或者：\usepackage{tcolorbox}\tcbuselibrary{breakable,skins}
\usepackage{multirow}
\usepackage{makecell}
\definecolor{colorcardbox}{RGB}{240, 248, 255}
\definecolor{colorcardborder}{RGB}{52, 52, 173}
\usepackage{listings}
\tcbuselibrary{listings,breakable}

\newtcolorbox{promptbox}[1][]{
  breakable,
  colback=colorcardbox,
  colframe=colorcardborder,
  arc=2pt,
  title={#1},
  fonttitle=\bfseries,
  left=5pt,
  right=5pt,
  top=5pt,
  bottom=5pt,
  before skip=1em,
  after skip=1em,
  fontupper=\small
}

\usepackage{wrapfig}
\usepackage{amsmath}
\usepackage{amssymb}
\usepackage{mathtools}
\usepackage{amsthm}

\usepackage[capitalize,noabbrev]{cleveref}
\usepackage{titletoc}
\usepackage{fancyhdr}

\theoremstyle{plain}

\theoremstyle{definition}

\theoremstyle{remark}

\newcommand{\company}[1]{}
\newcommand{\modelapi}[1]{{\texttt{#1}}}
\newcommand{\link}[1]{{\href{#1}{\texttt{Link}}}}
\usepackage[textsize=tiny]{todonotes}
\usepackage{dblfloatfix}
\newcommand{\ours}{\textsc{ZebraArena}\xspace}
\newcommand{\K}{\textsc{$K_{\text{ref}}$}\xspace}

\title{\ours: A Diagnostic Simulation Environment for Studying Reasoning–Action Coupling in Tool-Augmented LLMs}

\author{%
  Wanjia Zhao\textsuperscript{1}
  \quad
  Ludwig Schmidt\textsuperscript{1}
  \quad
  Yejin Choi\textsuperscript{1}
  \quad
  James Zou\textsuperscript{1}
  \\[0.4em]
  \textbf{Vidhisha Balachandran}\textsuperscript{2}
  \quad
  \textbf{Lingjiao Chen}\textsuperscript{2}
  \\[0.5em]
  \makebox[\textwidth][c]{%
    \textsuperscript{1}Stanford University
    \qquad
    \textsuperscript{2}Microsoft Research
  }
\vspace{-1.7em}
}

\fancypagestyle{firstpage}{
    \fancyhf{}

    \fancyfoot[L]{%
    \begin{minipage}{\textwidth}
        \rule{0.5\textwidth}{0.4pt}\\[0.25em]
        {\small
        \href{https://github.com/wanjiaZhao1203/zebraArena}{Github},
        \quad
        Corresponding to: 
        wanjiazh@cs.stanford.edu,
        lingjiaochen@microsoft.com
        }
    \end{minipage}
}
}
\begin{document}
\maketitle

\thispagestyle{firstpage}
\begin{abstract}
Tool-augmented large language models (LLMs) need to tightly couple multi-step reasoning with external actions, yet existing benchmarks often confound this interplay with complex environment dynamics, memorization effects, and dataset contamination. We introduce \ours, a procedurally generated \emph{diagnostic environment} for studying reasoning--action coupling in tool-augmented LLMs, with controllable difficulty and a knowledge-minimal design. Each task requires the agent to acquire missing information via targeted tool calls and integrate it through deductive reasoning, providing verifiable evaluation against unique ground-truth solutions and a constructive reference budget $\K$ for measuring tool-use efficiency. We show that \ours requires a combination of deep reasoning and accurate tool calling: even frontier reasoning models such as \texttt{Gemini 2.5 Pro} achieve only 60\% accuracy on the Large-size instances, and \texttt{GPT-5} uses 70--270\% more tool calls than $\K$ on medium puzzles, revealing a persistent gap between reasoning capability and efficient information acquisition. We release \ours to support future research on reasoning--action coupling in tool-augmented agents.
\end{abstract}
\section{Introduction}
\label{sec:intro}
Large language models (LLMs) augmented with external tools have become a central paradigm for building agents that combine reasoning with external information access~\citep{ schick2023toolformer, qin2023toolllm, gou2024tora, li2025torl, Jin2025SearchR1TL, song2025r1, chen2025researchlearningreason}. Such agents must not only produce correct answers, but also decide when to acquire additional information, which tools to invoke, and how to integrate retrieved evidence into multi-step reasoning~\citep{sun2024catpllmempoweringlarge,li2025flow,wang2025acting}. This interleaving of reasoning and action, exemplified by frameworks like ReAct~\citep{yao2023react}, is critical for accurate and reliable agent behavior.
Despite rapid progress, evaluating how tool-augmented LLMs coordinate information-seeking actions with deductive reasoning remains challenging. Many existing benchmarks~\citep{kwiatkowski2019natural,joshi2017triviaqa,yang2018hotpotqa,mallen2023not,press2022measuring,trivedi2022musique} primarily emphasize factual retrieval, where once the relevant information is obtained reasoning becomes shallow. In such settings, the agent's ``action'' often reduces to deciding whether the required knowledge is already known, rather than supporting sustained deductive reasoning. Another widely used class of benchmarks for studying tool use in RL focuses on math problem solving~\citep{he2024olympiadbench,wang2025acting,li2025torl,hendrycks2021measuring}, where agents frequently rely on a code interpreter for symbolic or numerical computation; however, such settings can bias the evaluation toward calculation and execution, and do not cleanly isolate the reasoning–action tradeoff under uncertainty.
At the other extreme, benchmarks that place agents in rich environments~\citep{Li2023APIBankAC,zhou2023webarena,xie2024travelplanner,jiang2024towards,qin2023toolllm, guo2025stabletoolbench,patil2024gorilla} (e.g., web navigation or embodied tasks) entangle reasoning with numerous confounds, including interface complexity, stochastic dynamics, and domain-specific heuristics. 
In all cases above, failures are hard to attribute to deficient reasoning, ineffective information acquisition, or environmental noise. These issues are compounded by reliance on long-standing public datasets, which raises contamination concerns, and by evaluation protocols that prioritize final task accuracy while offering limited support for analyzing \emph{when}, \emph{how}, and \emph{how efficiently} tools are used% More broadly, the structure of these benchmarks often lacks explicit notions of controllable uncertainty or information optimality
, making it difficult to support comprehensive, multi-dimensional evaluation of reasoning–action coupling.

To address these issues, we design an \emph{idealized simulation environment} with four goals: (i) \emph{knowledge-minimal}, so success depends on logical reasoning rather than domain facts; (ii) \emph{controllable}, with difficulty scalable along interpretable axes; (iii) \emph{contamination-resistant} by procedural construction of unlimited novel instances; and (iv) \emph{tightly coupled}, with explicit trade-offs between action cost and reasoning depth. Inspired by simplified testbeds and synthetic gym environments in reinforcement learning~\citep{menashe2018escape,parmar2024logicbench,terry2021pettingzoo,towers2024gymnasium,lin2025zebralogic}, we introduce \ours, a controlled simulation environment for studying reasoning--action coupling in tool-augmented LLMs under well-defined information constraints.

\begin{figure*}[t]
    \centering
    \includegraphics[width=\textwidth]{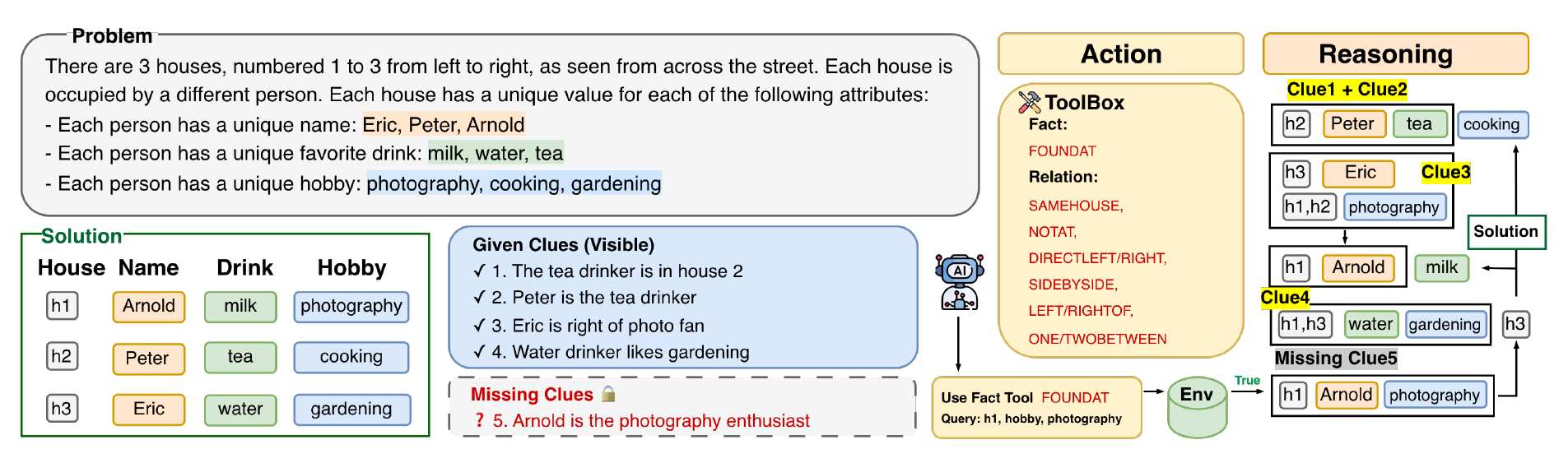}
    \caption{This \ours example has 3 houses $(N=3)$, 3 attributes $(M=3)$, and 5 clues $(K=5)$, with one withheld as a \emph{missing clue}. The \textbf{Background} defines attributes and uniqueness constraints, and the \textbf{Given Clues} provide visible constraints. The agent solves the puzzle by reasoning over the given clues and, when needed, querying the \textbf{ToolBox} to retrieve missing information from the environment. The final assignment is shown in the \textbf{Solution} grid.}
    \label{fig:overall}
\end{figure*}

\ours is derived from the classic Zebra (Einstein) puzzle~\citep{groza2021einstein, lin2025zebralogic}, a constraint satisfaction problem with a unique solution obtainable via pure logical reasoning, in which each of several houses must be assigned unique attributes (e.g., color, pet) consistent with a given clue set. We create partially observed instances by withholding a subset of logically necessary clues and exposing them only via tool queries, each returning a semantically well-defined piece of evidence. Solving an instance therefore requires agents to strategically acquire missing information and integrate it into deductive reasoning.
Difficulty is controlled by grid size and the number of missing clues, and the construction yields a per-instance reference budget $\K$ that anchors efficiency measurement. Unique, verifiable solutions further make evaluation deterministic.

Although \ours is based on synthetic puzzles, it captures a common structure in deployed agent settings: an agent must identify and request the missing information needed to make a reliable decision. In travel planning or calendar scheduling, users rarely state all preferences upfront, so agents must issue follow-up queries to narrow the feasible plan space; in API-based assistants, redundant calls inflate token cost and user burden. \ours abstracts this dynamic into a controlled, verifiable form, so that the diagnostics measured here (when to ask, what to ask, when to stop) are directly relevant to the design and selection of agents in such deployments. We thus view \ours as \emph{complementary to} realistic web/API benchmarks rather than a replacement, isolating reasoning--action coupling from interface noise and memorization.

Using \ours, we study tool-augmented LLM behavior under settings that vary task difficulty, query budgets, and query cost signals. This design enables joint assessment of accuracy and query efficiency, revealing a persistent gap between the constructive reference budget and practical tool usage. Across settings, we observe systematic strategy shifts under cost pressure and characteristic failure modes such as redundant querying loops, while certain information acquisition patterns remain unexpectedly robust at higher difficulties. Our key findings include: (1) GPT-5 achieves near-perfect accuracy ($\sim$99\%) on medium-difficulty puzzles, yet still uses 70--270\% more tool calls than the constructive reference budget; (2) Gemini-2.5-Flash requires over an order of magnitude more tokens per puzzle than GPT-5 ($\sim$20k vs.\ $\sim$1.2k); and (3) weaker models such as Llama-3.3-70B achieve only 12--24\% accuracy on medium puzzles, with high rates of insufficient information gathering.

Together, these results position \ours not only as an evaluation benchmark but as a diagnostic simulation environment for analyzing how tool-augmented LLM agents reason and adapt under well-defined constraints. In summary, our contributions are:
(1) \ours, a knowledge-minimal, contamination-resistant simulation with a constructive reference budget $\K$ for measuring tool-use overhead;
(2) a four-level (necessity $\to$ validity $\to$ utility $\to$ optimality) tool-use evaluation framework with efficiency-aware metrics beyond accuracy;
(3) diagnostic experiments across difficulty, query budgets, and pricing signals that reveal systematic gaps between reasoning capability and efficient information acquisition.

\section{Related Work}

\paragraph{Tool-Augmented Language Models}
Tool use in LLMs has been explored through diverse paradigms. Self-supervised approaches like Toolformer~\citep{schick2023toolformer} train models to autonomously decide when and how to invoke APIs, while ToolLLM~\citep{qin2023toolllm} and Gorilla~\citep{patil2024gorilla} focus on scaling to large tool repositories. For mathematical reasoning, ToRA~\citep{gou2024tora} integrates computation libraries and symbolic solvers, while TORL~\citep{li2025torl} incorporates code interpreters. The ReAct framework~\citep{yao2023react} established a prompting paradigm that interleaves reasoning traces with tool calls, inspiring subsequent work on cost-aware planning~\citep{sun2024catpllmempoweringlarge} and efficient action selection~\citep{wang2025acting}.
Evaluation of tool-augmented agents spans multiple settings. Multi-hop QA benchmarks~\citep{yang2018hotpotqa, trivedi2022musique, ho2020constructing} test retrieval-augmented reasoning but offer limited control over information availability. Web and API environments~\citep{zhou2023webarena, xie2024travelplanner, Li2023APIBankAC} provide realistic interaction but conflate reasoning errors with interface failures. Our work complements these efforts by providing a controlled setting where tool queries have well-defined semantics and a constructive reference query budget is known, enabling precise measurement of information acquisition efficiency.

\paragraph{Logic Puzzle Reasoning}
Logic puzzles have emerged as important benchmarks for evaluating LLM reasoning capabilities due to their knowledge-minimal nature and verifiable solutions. ZebraLogic~\citep{lin2025zebralogic} evaluates LLMs on zebra (Einstein) puzzles, demonstrating that performance degrades substantially as grid size increases. CLUTRR~\citep{sinha2019clutrr} tests compositional reasoning through kinship inference, revealing failures in systematic generalization. For suppositional reasoning, Knights and Knaves puzzles~\citep{mondorf2024truthquest} challenge models to handle truth-value assignments under hypothetical conditions. Several works augment LLMs with symbolic solvers: Logic-LM~\citep{pan2023logiclm} translates natural language into symbolic formulations for deterministic solving, while SatLM~\citep{ye2023satlm} leverages satisfiability solvers via declarative prompting. These benchmarks focus on fully-observed settings; our work introduces partial observability where agents must strategically acquire missing information.

%\subsection{Tool-Augmented Language Models Evaluation}
%An increasing number of work studies language models augmented with external tools, enabling agents to interleave reasoning and action for complex tasks. Representative approaches include ReAct-style prompting, tool-conditioned training, and function-calling agents. While these methods show strong empirical performance, evaluation typically relies on benchmarks that either emphasize factual retrieval or place agents in complex interactive environments.

%In retrieval-heavy benchmarks, reasoning becomes shallow once relevant information is obtained, making it difficult to disentangle reasoning ability from retrieval quality. Conversely, agentic environments introduce confounding factors such as interface complexity, stochastic transitions, and domain-specific heuristics, which obscure the analysis of reasoning--action coupling. As a result, existing evaluations provide limited insight in how agents acquire, use, and overuse information during reasoning.

\section{\ours}
\label{sec:formulation}
\subsection{Task Definition}
\paragraph{Zebra Puzzles} Zebra puzzles (logic grid puzzles) are canonical instances of \emph{Constraint Satisfaction Problems} (CSPs). We consider $N$ houses $\mathcal{H}=\{h_1,\ldots,h_N\}$ and $M$ attributes $\mathcal{A}=\{A_1,\ldots,A_M\}$ (e.g., \texttt{Name}, \texttt{Color}, \texttt{Food}), where each $A_j$ has values $D_j$ with $|D_j|=N$. An assignment $x : \mathcal{H} \times \mathcal{A} \to \bigcup_j D_j$ maps each $(h_i, A_j)$ to a value in $D_j$, with each $A_j$ defining a bijection between houses and $D_j$ (\emph{all-different constraint}). A puzzle is uniquely solvable when the clue set $\mathcal{C}_{\text{full}}$ admits exactly one such assignment, $|\{x \mid x \models \mathcal{C}_{\text{full}}\}|=1$. An instance is thus $P = (\mathcal{H}, \mathcal{A}, \mathcal{C}_{\text{full}}, S)$, where $S$ denotes this unique ground-truth solution.

We construct partially observed instances by sampling an incomplete clue set $\mathcal{C}_{0} \subsetneq \mathcal{C}_{\text{full}}$, which is insufficient for uniqueness: writing $\mathcal{X}(\mathcal{C}) \triangleq \{x \mid x \models \mathcal{C}\}$ for the feasible set under constraints $\mathcal{C}$, we have $|\mathcal{X}(\mathcal{C}_0)| > 1$. As the agent acquires additional constraints, the feasible set contracts monotonically, $|\mathcal{X}(\mathcal{C}_{t+1})| \le |\mathcal{X}(\mathcal{C}_{t})|$, and collapses to $\{S\}$ once enough constraints are recovered. Solving therefore requires querying the environment for targeted information and integrating it with $\mathcal{C}_{0}$ to converge to $S$.

\subsection{Environment and Tools}
The puzzle environment is a \emph{rule-based oracle} derived from the ground-truth solution $S$, supporting two families of queries:

\emph{\textbf{Fact queries.}} A fact query tests a specific house--attribute assignment, $\{ \text{type}:\text{"fact"}, \text{house}:h_i, \text{attr}:A_j, \text{value}:v \}$, returning a Boolean for whether $(h_i, A_j)\mapsto v$ in $S$.

\emph{\textbf{Relation queries.}} A relation query checks whether a logical relation holds in $S$, including positional (\texttt{left\_of}, \texttt{right\_of}, \texttt{adjacent}), distance (\texttt{distance\_k}, \texttt{exactly\_k\_left\_of}), and structural predicates (\texttt{same\_house}, \texttt{at\_index}, \texttt{is\_first}, \texttt{is\_last}).

\paragraph{Schema validation and deterministic response.}
All agent queries are validated against JSON schemas (with variants per environment type: normal / fact-only / relation-only) and canonicalized via case- and whitespace-insensitive normalization, so semantically equivalent queries are treated identically. Validated queries are executed by the rule-based server, which returns exact, noise-free, reproducible Boolean answers (or an explicit error for invalid calls). This deterministic semantics isolates reasoning--action behavior from retrieval noise and environment stochasticity. After $t$ queries, the cumulative constraint set is $\mathcal{C}_t=\mathcal{C}_0 \cup \{\phi(q_i,a_i)\}_{i\le t}$, where $\phi(q,a)$ is the constraint implied by answer $a$; the feasible set is monotone non-increasing in $t$. Figure~\ref{fig:exp1_overall} illustrates a puzzle with its reasoning chain and tool calls.

\subsection{Controllable Complexity}
Difficulty in \ours is controlled along two axes. The first is grid size $(N, M)$: under all-different constraints, the unconstrained solution space has size $(N!)^{M}$, growing factorially with grid size and providing a model-independent measure of intrinsic complexity. The second is the number of missing clues, which increases initial uncertainty while keeping the ground-truth solution fixed and requires agents to acquire missing constraints via tool queries.

\paragraph{Constructive reference budget.}
Let $\K$ be the number of atomic clues masked during construction. Since each masked clue is normalized into a predicate family supported by the Normal tool schema, it has a corresponding admissible query, and each instance admits a direct recovery policy obtaining all missing clues in exactly $\K$ queries. We use $\K$ as an achievable reference budget for measuring tool-use overhead.
$\K$ is not the globally optimal adaptive query complexity---some instances may be solved with fewer queries via more informative predicates---but our metric only relies on the constructive guarantee that a $\K$-query recovery policy exists. As a calibration check, Appendix~\ref{app:oracle_calibration} reports an oracle decision-tree baseline whose expected query count is close to $\K$. Thus, models using more than $\K$ queries incur overhead relative to a guaranteed feasible strategy, not an unattainable oracle.

\subsection{Evaluation Metrics}
We evaluate models on \ours under the \emph{missing-clues} setting using a four-level diagnostic framework that progressively constrains \emph{when} tools are invoked, \emph{how} calls are executed, and \emph{how well} they advance progress, decomposing tool-augmented reasoning into task planning, tool selection, tool calling, and response generation.

    \emph{\textbf{1. Necessity: Should the agent call a tool at this step?}}
    This level evaluates task planning decisions. A step is correct if the agent calls a tool only when current information is insufficient, and directly proposes a solution when sufficient information is already available. Redundant or premature tool calls are reflected in increased \emph{total reasoning steps} and \emph{redundant tool calls}, and contribute to overhead relative to the reference query budget $\K$.

  \emph{\textbf{2. Validity (Syntax \& Semantics): Is the tool call well-formed and permissible?}}
This level checks whether a tool call (i) conforms to the expected JSON schema and (ii) references valid query types, relations, and entities supported by the environment. Invalid calls—including format errors or impermissible/undefined actions—are reflected by the gap between \emph{total tool calls} and \emph{valid tool calls}, and are excluded from subsequent utility-based analysis.

    \emph{\textbf{3. Utility: Does the tool call meaningfully reduce uncertainty?}}
    Writing $|\mathcal{X}(\mathcal{C}_t)|$ for the number of feasible solutions consistent with the constraints accumulated by step $t$, we measure per-step information gain $\mathrm{IG}_t = \log |\mathcal{X}(\mathcal{C}_{t-1})| - \log |\mathcal{X}(\mathcal{C}_t)|$ and an effectiveness indicator $E_t = \mathbb{1}[|\mathcal{X}(\mathcal{C}_t)| < |\mathcal{X}(\mathcal{C}_{t-1})|]$ capturing whether a query reduces uncertainty.

   \emph{\textbf{4. Optimality: Is information acquired near-optimally?}}
    Across $T$ tool calls, we compute the effectiveness rate $\mathrm{EffRate}=\tfrac{1}{T}\sum_t E_t$ and mean information gain $\overline{\mathrm{IG}}=\tfrac{1}{T}\sum_t \mathrm{IG}_t$. To assess redundancy relative to the constructive reference budget, we define the \emph{inefficiency ratio}
    \[ \small
    \mathrm{IR} = \frac{T}{\K},
    \qquad
    \mathrm{IR}_{\text{eff}} = \frac{\sum_{t=1}^T E_t}{\K},
    \]
    which measure total and effective query overhead, respectively. $\mathrm{IR}$ is reported relative to $\K$, not the unknown globally optimal adaptive policy.

\subsection{Dataset Generation and Statistics}
Our dataset extends ZebraLogic~\citep{lin2025zebralogic}, which provides text-based zebra puzzles with unique solutions. 
We construct a \emph{missing-clues} variant by masking a subset of clues, and evaluate across Small/Medium/Large sizes with 1--6 missing clues. 
The resulting benchmark contains over 2000 instances; construction details and distributions are deferred to Appendix~\ref{app:dataset_curation_details}.

\begin{figure*}[t]
    \centering
    \includegraphics[width=\textwidth]{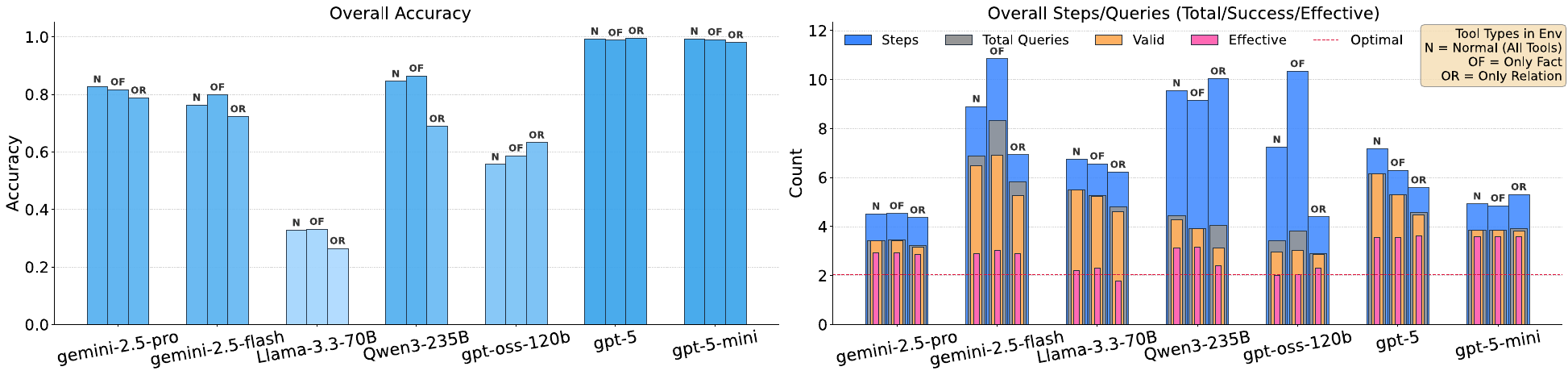}
    \caption{Overall accuracy and hierarchical tool-use diagnostics on \ours.\textit{ Left:} accuracy for each model under three tool environment types (Normal / Only-Fact / Only-Relation). \textit{Right:} Interaction counts aligned with our 4-level evaluation metrics: Steps, Total Queries, Valid Queries, Effective Queries. Gaps between bars isolate failures at each stage, with the red dashed line marking the reference budget $\K$.}
    \label{fig:exp1_overall}
\end{figure*}

\section{Experiments}

\subsection{Setup}
This experiment measures baseline performance across complexity levels and tool settings. We vary two factors:
\textbf{Complexity:} Puzzle size (Small, Medium, Large) and number of missing clues ($\K \in \{1,2,3,4,5,6\}$), both of which exponentially increase the solution space.
\textbf{Tool availability:} Three settings: \textsc{Normal} with both fact and relation queries, \textsc{OnlyFact} with fact queries only, and \textsc{OnlyRelation} with relation queries only.

\subsection{Main Results}
Figure~\ref{fig:exp1_overall} reports accuracy, reasoning steps, and tool calls across models, with additional results in Appendix~\ref{app:exp1}. Ideally, valid tool calls equal total tool calls and match $\K$, with total steps $\K+1$ for the final answer. Stronger reasoning models (e.g., \texttt{gpt-5}) come closer to this pattern, making mostly valid queries with few redundant steps, whereas smaller chat models often substitute excessive tool use for structured reasoning. This suggests explicit deliberation helps allocate queries adaptively. However, no model reaches the reference budget: overhead remains through redundant queries, invalid calls, or failure to stop within $\K{+}1$ steps.
\begin{figure*}[t]
    \centering
    \includegraphics[width=\textwidth]{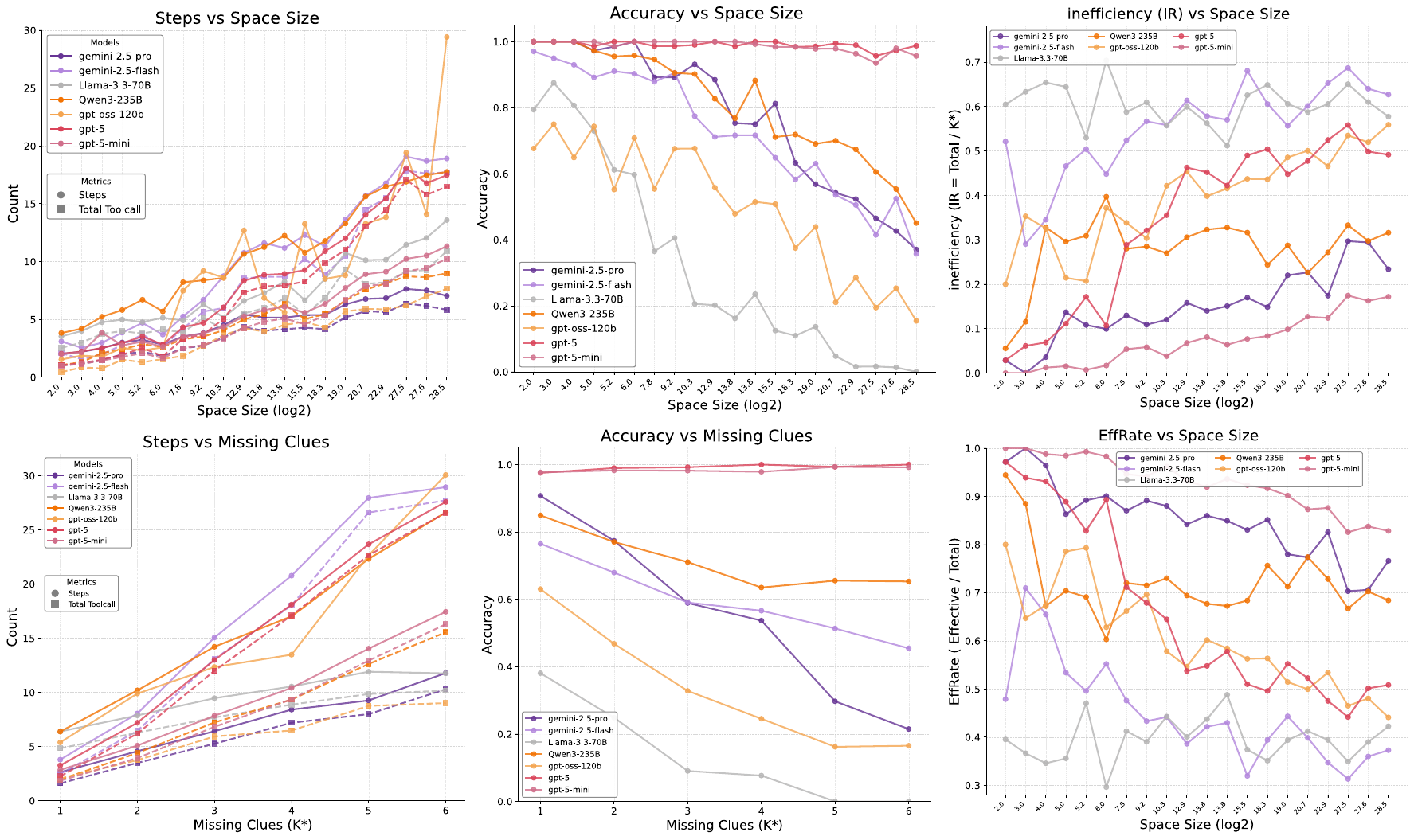}
    \caption{Scaling behavior of tool-augmented reasoning on \ours. \textit{Left:} Interaction cost as task complexity increases, measured by total steps and  total tool calls, 
     shown as a function of search-space size (top) and number of missing clues $\K$ (bottom). \textit{Middle:} Accuracy under the same scaling axes. \textit{Right:} tool-use efficiency (inefficiency ratio and effectiveness rate), highlighting the widening overhead relative to the reference budget as uncertainty grows.}
    \label{fig:exp1_scaling}
\end{figure*}

\subsection{Scaling Puzzle Complexity}
\begin{wraptable}{r}{0.5\textwidth}
\vspace{-2em}
\centering
\small
\caption{Average token usage per puzzle, per message, and mean information gain (IGmean), across models and environment types on Medium Size.}
\label{tab:exp1_token}
\resizebox{\linewidth}{!}{%
\begin{tabular}{llccc}
\toprule
\textbf{Model} & \textbf{Environment} & \textbf{AvgPerPuzzle} & \textbf{AvgPerMessage} & \textbf{IGmean} \\
\midrule
GPT-5 & normal         & 1{,}196  & 116  & 0.5518 \\
      & only\_fact     & 1{,}088  & 127  & 0.5889 \\
      & only\_relation & 1{,}008  & 135  & 0.6321 \\
\midrule
GPT-5-mini & normal         & 1{,}524  & 249  & 0.6906 \\
          & only\_fact     & 1{,}525  & 251  & 0.6724 \\
          & only\_relation & 1{,}424  & 211  & 0.7014 \\
\midrule
Gemini-2.5-flash & normal         & 19{,}898 & 1{,}733 & 0.3259 \\
                 & only\_fact     & 25{,}396 & 1{,}857 & 0.3051 \\
                 & only\_relation & 20{,}757 & 2{,}392 & 0.4131 \\
\midrule
Gemini-2.5-pro   & normal         & 15{,}053 & 2{,}810 & 0.5697 \\
                 & only\_fact     & 14{,}865 & 2{,}747 & 0.6202 \\
                 & only\_relation & 16{,}676 & 3{,}237 & 0.6260 \\
\midrule
Qwen3-235B & normal         & 9{,}508  & 838  & 0.5297 \\
          & only\_fact     & 9{,}503  & 868  & 0.6110 \\
          & only\_relation & 12{,}029 & 1{,}025 & 0.4756 \\
\midrule
Llama-3.3-70B & normal         & 3{,}304 & 427 & 0.2632 \\
             & only\_fact     & 3{,}490 & 458 & 0.2385 \\
             & only\_relation & 3{,}949 & 542 & 0.3007 \\
\midrule
GPT-OSS-120b & normal         & 2{,}374 & 263 & 0.3079 \\
            & only\_fact     & 2{,}402 & 170 & 0.2965 \\
            & only\_relation & 695     & 125 & 0.4605 \\
\bottomrule
\end{tabular}
}
\vspace{-3em}
\end{wraptable}
Figure~\ref{fig:exp1_scaling} examines how performance and tool use scale with puzzle complexity in the missing-clues setting. We vary (i) the combinatorial difficulty, measured by the solution-space size (space size, $\log_2 |\mathcal{S}|$), and (ii) the information deficit, measured by the number of missing clues $\K$. Across models, increasing either axis lowers accuracy and increases interaction cost, while also reducing query efficiency.

\paragraph{Scaling by search space.}
As the solution space grows, accuracy decreases while both reasoning steps and total tool calls increase, indicating that larger combinatorial spaces require longer interaction horizons to resolve ambiguity.
Query efficiency also decreases: IR rises with space size, meaning models spend progressively more queries relative to the reference budget $\K$, and EffRate declines, indicating that fewer queries effectively reduce the candidate space. This effect is strongest for weaker models, which exhibit faster growth in redundant or ineffective queries, whereas stronger models remain closer to the reference budget but still incur nontrivial overhead as complexity scales.

\paragraph{Scaling by missing clues.}
At fixed puzzle size, increasing $\K$ reduces accuracy (bottom-middle) while monotonically increasing steps and tool calls (bottom-left), as more withheld constraints must be recovered. Interaction growth exceeds the increase in $\K$, indicating compounding overhead under partial observability: models not only issue more queries, but also spend more reasoning effort selecting queries and integrating answers.

Overall, increasing either space size or information deficit stresses both reasoning and action, pushing models further from the reference budget.

\begin{table*}[t]
\centering
\caption{Insufficiency analysis across models and puzzle settings. We report the fraction of puzzles that are \emph{insufficient}, meaning the agent makes fewer than $\K$ valid tool calls and thus fails to gather enough information to determine a unique solution. For each missing-clue level, we show two numbers: the insufficient rate (\%) and the accuracy gap $\Delta$Acc between the remaining (non-insufficient) and insufficient subsets (Rem--Insuf).}
\resizebox{\linewidth}{!}{%
\begin{tabular}{l|cc|cc|cc|cc|cc|cc|cc|cc|cc|cc|cc|cc}
\toprule
\multirow{2}{*}{Model}
& \multicolumn{4}{c|}{\textbf{Small}}
& \multicolumn{8}{c|}{\textbf{Medium}}
& \multicolumn{12}{c}{\textbf{Large}} \\
& \multicolumn{2}{c}{Missing1} & \multicolumn{2}{c|}{Missing2}
& \multicolumn{2}{c}{Missing1} & \multicolumn{2}{c}{Missing2} & \multicolumn{2}{c}{Missing3} & \multicolumn{2}{c|}{Missing4}
& \multicolumn{2}{c}{Missing1} & \multicolumn{2}{c}{Missing2} & \multicolumn{2}{c}{Missing3} & \multicolumn{2}{c}{Missing4} & \multicolumn{2}{c}{Missing5} & \multicolumn{2}{c}{Missing6} \\
& Insuf & $\Delta$Acc & Insuf & $\Delta$Acc
& Insuf & $\Delta$Acc & Insuf & $\Delta$Acc & Insuf & $\Delta$Acc & Insuf & $\Delta$Acc
& Insuf & $\Delta$Acc & Insuf & $\Delta$Acc & Insuf & $\Delta$Acc & Insuf & $\Delta$Acc & Insuf & $\Delta$Acc & Insuf & $\Delta$Acc \\
\midrule

GPT-5
& 0.3 & -1.0 & 0.5 & +0.0
& 3.8 & +32.9 & 2.5 & +33.3 & 0.9 & +50.0 & 0.0 & --
& 5.9 & +63.9 & 2.0 & +99.3 & 0.7 & +99.3 & 0.0 & -- & 0.0& -- & 0.0 & -- \\

GPT-5-mini
& 0.0 & --   & 2.6 & -0.5
& 0.8 & +49.6 & 2.1 & +20.0 & 0.9 & +49.1 & 0.6 & -1.2
& 7.2 & +58.7 & 2.0 & +62.7 & 0.7 & -2.6  & 0.7 & +97.4 & 0.0 & -- & 0.0 & -- \\
Gemini-2.5-Flash
& 9.3 & +46.4 & 9.5 & +55.9
& 23.3 & +35.2 & 23.0 & +52.4 & 13.9 & +67.3 & 12.7 & +65.1
& 23.7 & +31.7 & 26.2 & +50.0 & 23.5 & +48.3 & 21.6 & +58.3 & 13.5 & +53.6 & 21.5 & +57.9 \\

Gemini-2.5-Pro
& 1.0 & +32.7 & 3.7 & +65.4
& 10.2 & +40.1 & 13.1 & +51.6 & 12.0 & +61.9 & 16.8 & +75.0
& 15.1 & +34.6 & 23.5 & +71.4 & 28.1 & +51.2 & 24.2 & +52.6 & 33.1 & +44.4 & 37.2 & +34.2 \\

Qwen3-235B
& 2.6 & +60.5 & 1.6 & +97.3
& 16.9 & +47.8 & 8.9 & +59.4 & 4.7 & +66.1 & 13.9 & +73.5
& 25.7 & +36.1 & 24.2 & +51.1 & 13.1 & +52.4 & 13.7 & +57.4 & 5.4 & +69.3 & 8.3 & +71.2 \\

Llama-3.3-70B
& 6.6 & +39.1 & 5.3 & +46.1
& 16.9 & +13.5 & 10.2 & +4.0 & 4.7 & +14.3 & 7.5 & +12.5
& 17.8 & +4.0 & 15.7 & +3.1 & 24.2 & +2.6 & 12.4 & +3.7 & 10.8 & +0.0 & 16.5 & +0.0 \\

GPT-OSS-120B
& 46.0 & +45.9 & 44.2 & +78.4
& 29.2 & +38.7 & 30.1 & +59.4 & 30.8 & +54.5 & 39.9 & +49.0
& 33.6 & +26.0 & 40.5 & +24.9 & 39.9 & +28.8 & 40.5 & +26.4 & 38.5 & +23.5 & 49.6 & +32.8 \\

\bottomrule
\end{tabular}
}
\label{tab:main_insuf}
\end{table*}

\subsection{Insufficient Query Analysis}
Table~\ref{tab:main_insuf} reports the insufficient rate, defined as cases where the number of valid tool calls falls below $\K$. Overall, insufficient rates increase monotonically with problem difficulty, rising from Small to Medium/Large and from fewer to more missing clues. This trend indicates that harder settings more often prevent agents from collecting even a minimal amount of decisive information needed to resolve the puzzle. Notably, GPT-5 and GPT-5-mini maintain near-zero insufficient rates across all evaluated settings, whereas weaker models---most prominently GPT-OSS-120B and Llama-3.3-70B---exhibit substantial insufficiency even at moderate difficulty levels. Notably, GPT-OSS-120B shows very high insufficient rates on Small puzzles, indicating an early-stopping pattern: when the search space is relatively small, it often commits to solving early rather than systematically using tools to verify and narrow down candidates. These results suggest that a significant fraction of failures arises not from downstream reasoning errors alone, but from early breakdowns in information acquisition under uncertainty.

\subsection{Tool Usage Across Three Environment Types}
We analyze tool use by jointly considering accuracy, query efficiency, and token consumption across fact-only, relation-only, and mixed environments (Figure~\ref{fig:exp1_overall}, Table~\ref{tab:exp1_token}). 

Under the fact-only environment, models tend to achieve higher accuracy on harder instances since fact queries provide decisive, low-level information that directly resolves individual unknowns; however, token consumption is model-dependent---weaker models (e.g., Gemini-2.5-Flash, Llama-3.3-70B) incur substantially higher costs due to increased tool calls, while capable models like GPT-5 leverage fine-grained queries efficiently without interaction overhead.

Under the relation-only environment, models issue fewer tool calls with fewer interaction turns, but each message consumes more tokens due to the reasoning required to interpret relational constraints; this heavier reliance on internal deduction also makes relation-only strategies more sensitive to early errors, reducing stability as uncertainty grows.

Mixed settings provide flexibility, reducing redundant queries while maintaining accuracy. These results expose an efficiency--robustness trade-off: fact queries resolve uncertainty directly but lengthen interaction, whereas relation queries shorten it at the cost of heavier per-step reasoning and lower stability---illustrating \ours's ability to evaluate cost and stability beyond accuracy alone.

\begin{table}[!htbp]
\centering
\caption{Error analysis across missing-clue levels and puzzle sizes. \emph{Reason}/\emph{Insuff Info Err}: incorrect answer with $|S(\mathcal C_t)|{=}1$ vs.\ $>1$; \emph{Acc @ Singleton/Ambiguous}: conditional accuracy by final feasible-set size; \emph{Over-Query}: runs that keep querying after $|S(\mathcal C_t)|{=}1$.}
\label{tab:error_analysis}
\small
\setlength{\tabcolsep}{4pt}
\begin{tabular}{ll|cc|c|cc|c}
\toprule
Setting & Size
& \makecell{Reason\\Err (\%) $\downarrow$}
& \makecell{Insuff Info\\Err (\%) $\downarrow$}
& \makecell{Insuff /\\Reason}
& \makecell{Acc @\\Singleton (\%) $\uparrow$}
& \makecell{Acc @\\Ambiguous (\%) $\uparrow$}
& \makecell{Over-Query\\(\%) $\downarrow$} \\
\midrule
\multirow{3}{*}{Missing1}
 & Small  & 3.2 &  7.6 &  2.4$\times$ & 96.3 & 43.7 & 15.4 \\
 & Medium & 4.7 & 20.2 &  4.3$\times$ & 93.2 & 33.1 & 13.0 \\
 & Large  & 6.5 & 31.8 &  4.9$\times$ & 88.5 & 27.2 & 14.3 \\
\midrule
\multirow{3}{*}{Missing2}
 & Small  & 3.2 & 13.0 &  4.1$\times$ & 96.2 & 23.8 & 18.3 \\
 & Medium & 3.0 & 27.5 &  9.2$\times$ & 95.5 & 19.4 & 14.5 \\
 & Large  & 3.4 & 42.7 & 12.6$\times$ & 93.3 & 13.9 & 11.8 \\
\midrule
\multirow{3}{*}{Missing3}
 & Small  & 2.2 &  6.7 &  3.0$\times$ & 97.4 & 50.0 & 62.2 \\
 & Medium & 2.0 & 30.3 & 15.2$\times$ & 97.0 & 13.6 & 18.8 \\
 & Large  & 2.7 & 46.0 & 17.0$\times$ & 94.5 &  9.0 & 15.6 \\
\bottomrule
\end{tabular}
\end{table}

\subsection{Error Analysis}
\ours decomposes model failures along two axes: whether the answer is correct and whether the final feasible set $\mathcal{X}(\mathcal{C}_t)$ is a singleton. This yields four categories: \emph{Full Solve} (correct, singleton), \emph{Lucky Guess} (correct, ambiguous), \emph{Reasoning Error} (incorrect, singleton), and \emph{Insufficient-Information Error} (incorrect, ambiguous). The two error types separate failures of downstream reasoning from failures of information acquisition and stopping.

Across models, we find that the dominant failure mode is not downstream reasoning, but insufficient information acquisition and incorrect stopping. As shown in Table~\ref{tab:error_analysis}, reasoning errors are consistently rare ($\sim 2$--$7\%$), while insufficient-information errors grow sharply with puzzle size and missing-clue level, reaching $46.0\%$ on Large puzzles---an Insuff/Reason ratio up to $17\times$. Conditional accuracy reflects the same pattern: $88$--$97\%$ once the feasible set is a singleton, but only $9$--$50\%$ when residual ambiguity remains. We also observe nontrivial over-querying (up to $62.2\%$ on Missing3 Small), so models do not simply stop too early---they sometimes keep querying after sufficient information is already obtained. Together, this demonstrates the diagnostic granularity of \ours for separating query selection, integration, and stopping.

\subsection{Budget Constraints}
This experiment studies how resource constraints shape agent behavior and performance. We inform agents of their query budget in the prompt and evaluate whether they can solve puzzles within the allocation.
We define three budget levels based on the constructive reference budget $\K$ and empirical tool-use statistics from Exp1:
\textbf{Tight:} $\text{Budget}=\K$.
\textbf{Normal:} $\text{Budget}\approx$ the mean tool-call count in Exp1 under the unconstrained Baseline.
\textbf{Relaxed:} a larger budget that adds slack beyond Normal (see Appendix~\ref{app:exp2} for detailed values).
We adopt \textbf{soft constraints}: agents are shown the budget and remaining queries after each turn, but are not forcibly stopped when exceeding it, enabling analysis of both adherence and performance under pressure.

Table~\ref{tab:exp2} summarizes the impact of budget constraints on task performance and tool use. Overall, explicitly specifying a budget leads agents to reduce tool-call counts relative to the unconstrained Baseline, indicating that agents do respond to cost signals even under soft constraints without hard termination.

Tight budgets ($\K$), though sufficient under the construction, cause substantial accuracy drops as agents spend early queries on non-decisive information. Normal budgets provide better trade-offs by suppressing redundant querying while preserving enough acquisition to maintain accuracy. Relaxed budgets help some models (e.g., Gemini-2.5-Pro) by translating the extra slack into more effective tool calls, but fail to recover Baseline accuracy for others and can even degrade performance on harder Medium puzzles. We refer to this pattern---agents overweighting cost and reducing tool use even when ample budget remains, failing to convert the available slack into uncertainty reduction---as \emph{budget anxiety}.

Overall, explicit budgets serve as a controllable intervention on tool use and reveal a clear feasibility frontier: $\K$ is brittle in practice, moderate slack improves robustness, and additional slack yields diminishing returns. This positions \ours as a diagnostic environment for studying how reasoning--action coupling degrades under resource constraints, and motivates future work on \emph{multi-objective optimization} that jointly optimizes task success, tool efficiency, and reliability rather than treating cost control as a purely external constraint.

\subsection{Pricing Signals}
\label{price}
This experiment examines whether agents can adapt their tool selection strategies in response to economic signals. We assign virtual token costs to each query type and instruct agents to minimize total cost while solving accurately.
We define pricing conditions with varying token price and cost ratios as shown in Appendix~\ref{app:exp3_setup}.
After each turn, agents are provided with cumulative resource usage,
\texttt{[Token usage: X reasoning + Y tools = Z total]}.
Our evaluation tracks the number of each type of tool query to quantify strategy shifts induced by price signals.

\subsubsection{Results and Analysis}

As shown in Table~\ref{tab:exp3_gemini2.5flash} and Table~\ref{tab:exp3_Qwen3_235B}, pricing in general reliably shifts the mix of tool calls: when one query type becomes cheaper, agents tend to substitute toward it, and when it becomes more expensive, they reduce its usage, indicating sensitivity to explicit cost cues. However, the performance impact is model- and difficulty-dependent. In general, moderate price perturbations often change behavior without substantially improving accuracy, and extreme asymmetry can even hurt by discouraging a query type that is still necessary for resolving uncertainty, especially as missing clues increase. Notably, Qwen3-235B exhibits more stable accuracy across pricing conditions while adjusting query allocation, whereas Gemini-2.5-Flash shows larger accuracy swings and higher variance, suggesting that cost-aware tool selection remains imperfect and can induce conservative or imbalanced information acquisition rather than consistently lower-cost, high-accuracy strategies.

\section{Conclusion and Future Work}
We introduced \ours, a controllable simulation environment for studying reasoning--action coupling in tool-augmented LLM agents. \ours formulates partially observed Zebra puzzles with unique, verifiable solutions and an explicit constructive reference budget $\K$, supporting multi-dimensional evaluation of accuracy, efficiency, and tool-use behavior. Our experiments reveal a consistent overhead relative to $\K$ and systematic trade-offs between robustness and resource use under different tool regimes and budget/cost constraints.

These diagnostics translate into actionable insights for two audiences. For \textbf{model providers}, \ours pinpoints which capability needs improvement: GPT-5 reaches near-perfect accuracy on medium puzzles yet uses 70--270\% more tool calls than $\K$, showing that strong reasoning does not imply efficient information acquisition; weaker models such as GPT-OSS-120B fail predominantly through insufficient-information errors (up to $46\%$ on Large puzzles), pointing to a different bottleneck in stopping and uncertainty management. For \textbf{model deployers}, the pricing analysis (Sec.~\ref{price}) shows that models with similar headline capability behave quite differently under deployment constraints---Qwen3-235B remains stable while Gemini-2.5-Flash shows large accuracy swings---informing model selection under cost or interaction-budget constraints.

Future work includes using \ours as a testbed for agents that explicitly model uncertainty and information value, extending the environment to richer cost and noise settings, transferring the same simulation design to other reasoning domains, and using \ours as a training environment for multi-objective optimization that jointly optimizes task success, tool efficiency, and reliability under changing constraints.

\newpage
\bibliography{neurips}
\bibliographystyle{neurips}

%%%%%%%%%%%%%%%%%%%%%%%%%%%%%%%%%%%%%%%%%%%%%%%%%%%%%%%%%%%%
\newpage
\appendix
%%%%%%%%%%%%%%%%%%%%%%%%%%%%%%%%%%%%%%%%%
\hrule height 4pt
\vskip 0.25in
\vskip -\parskip
%%%%%%%%%%%%%%%%%%%%%%%%%%%%%%%%%%%%%%%%%
\vbox{
    \centering
    \LARGE 
    \textbf{Supplementary Materials for \\  
   ZebraArena: A Diagnostic Simulation Environment for Studying Reasoning–Action Coupling in Tool-Augmented LLMs}
}
%%%%%%%%%%%%%%%%%%%%%%%%%%%%%%%%%%%%%%%%%
\vskip 0.29in
\vskip -\parskip
\hrule height 1pt
%%%%%%%%%%%%%%%%%%%%%%%%%%%%%%%%%%%%%%
\renewcommand*\footnoterule{} % remove the separator line
\newcommand\blfootnote[1]{%
  \begingroup
  \renewcommand\thefootnote{}\footnote{#1}%
  \addtocounter{footnote}{-1}%
  \endgroup
}
%%%%%%%%%%%%%%%%%%%%%%%%%%%%%%%%%%%%%%%%%
\makeatletter
\def\addcontentsline#1#2#3{%
  \addtocontents{#1}{\protect\contentsline{#2}{#3}{\thepage}{\@currentHref}}}
\makeatother

%%%%%%%%%%%%%%%%%%%%%%%%%%%%%%%%%%%%%%%%%
% Create a new ToC for appendix only
% \section*{Appendix Contents}
\setcounter{tocdepth}{2}
\renewcommand{\contentsname}{Appendix Contents}
\startcontents[appendix]  % Requires the 'titletoc' package
\printcontents[appendix]{}{1}{\setcounter{tocdepth}{2}}
%%%%%%%%%%%%%%%%%%%%%%%%%%%%%%%%%%%%%%%%%

\clearpage

\section{Dataset Curation Details}
\label{app:dataset_curation_details}

\subsection{Data Source and Collection}
Our dataset is constructed by extending the ZebraLogic benchmark~\citep{lin2025zebralogic}, which consists of text-based Zebra (logic grid) puzzles with logically sufficient clue sets and unique solutions. We follow the original problem setting and solution structure, but transform the data into a form suitable for systematic analysis of search-space dynamics and tool-augmented reasoning.

\paragraph{Clue normalization and symbolic representation.}
We normalize each natural-language clue into a structured, symbolic representation by mapping each clue to a predefined clue type (FOUNDAT, SAMEHOUSE, NOTAT, DIRECTLEFT/RIGHT, SIDEBYSIDE, LEFT/RIGHTOF, ONE/TWOBETWEEN). Each parsed clue is then represented as an atomic predicate with explicit operands (attribute--value entities) and a relation operator. This typed symbolic form supports downstream search-space computation and exact constraint checking. Concretely, a textual clue such as \emph{``The Dragonfruit smoothie lover is Eric.''} is mapped to a structured constraint of the form:   
\begin{verbatim}
{
  "id": "c1",
  "parsed": {
    "lhs": {"attr": "Smoothie", "value": "dragonfruit"},
    "rel": "same_house",
    "rhs": {"attr": "Name", "value": "eric"}
  },
  "type": "relation"
}
\end{verbatim}
This atomized representation makes all constraints explicit and compositional, enabling direct reasoning over constraint satisfaction and facilitating downstream computation of the feasible solution space.

\paragraph{Search space computation and filtering.}
Based on the symbolic constraints, we implement an exact solver that enumerates the feasible solution space induced by a given clue set. Using this solver, we filter out instances that do not admit a unique solution under the full clue set, ensuring consistency with the standard Zebra puzzle formulation and the ZebraLogic benchmark. This step guarantees that every retained instance has a well-defined ground-truth solution and a finite, computable search space.

\paragraph{Construction of missing-clues instances.}
To construct \textsc{ZebraArena}, we further transform each uniquely solvable instance into a family of partially observed problems. For each puzzle, we randomly mask a subset of clues, withholding between one and half of the original clues. The remaining clues constitute the initial constraint set $\mathcal{C}_0$, which is insufficient to uniquely determine the solution. For each masked instance, we compute the initial feasible solution count induced by $\mathcal{C}_0$, which serves as the starting uncertainty for the agent. The withheld clues are made accessible only through tool queries, forming the basis of the interactive missing-clues setting.

\subsection{Alternative environment construction.}
Beyond curating instances from ZebraLogic, \textsc{ZebraArena} can also be instantiated by directly adopting the ZebraLogic procedural generator to sample new puzzles on demand, and then applying the same missing-clues transformation. In this mode, key parameters—including grid size $(N,M)$, clue-set size, and the number of missing clues are fully user-configurable, yielding a highly controllable and flexible simulation environment that supports scalable instance generation without reliance on fixed datasets.

\subsection{Dataset Distribution}

Table~\ref{tab:dataset_distribution} summarizes the number of instances for each puzzle size and missing-clue level used in our experiments.

\begin{table}[htbp]
\centering
\caption{Dataset distribution across puzzle sizes and numbers of missing clues. ``--'' indicates settings not included for that size.}
\begin{tabular}{c|cccccc}
\toprule
\textbf{Size} & \multicolumn{6}{c}{\textbf{Missing Clues ($\K$)}} \\
 & 1 & 2 & 3 & 4 & 5 & 6 \\
\midrule
Small  & 302 & 190 & --  & --  & --  & --  \\
Medium & 236 & 236 & 234 & 173 & --  & --  \\
Large  & 152 & 153 & 153 & 153 & 148 & 121 \\
\bottomrule
\end{tabular}
\label{tab:dataset_distribution}
\end{table}

\section{Experimental Setups}
\label{app:exp_setups}
\subsection{Model Setup}
We evaluate 7 LLMs with different reasoning levels, as listed in Table \ref{app:table_model_and_params}. Each model is accessed via its official API using standardized decoding parameters. 
 \begin{table}[htbp]
\centering
\caption{List of LLMs evaluated in our experiments.}
\setlength{\tabcolsep}{5pt}
\renewcommand{\arraystretch}{1.25}

\newcounter{rownumber}
\setcounter{rownumber}{0}

% \begin{adjustbox}{max width=\textwidth}
\begin{tabular}{rlll}
\toprule
\# & \textbf{Model}  & \textbf{Model Engine Name} & \textbf{Source} \\
\midrule

\stepcounter{rownumber}\arabic{rownumber} & Gemini-2.5-Flash~\cite{google2025gemini2.5flash} & \modelapi{gemini-2.5-flash} & \link{https://cloud.google.com/vertex-ai/generative-ai/docs/models/gemini/2-5-flash} \\
\stepcounter{rownumber}\arabic{rownumber} & Gemini-2.5-Pro~\cite{google2025gemini2.5pro} & \modelapi{gemini-2.5-pro} & \link{https://cloud.google.com/vertex-ai/generative-ai/docs/models/gemini/2-5-pro} \\
\stepcounter{rownumber}\arabic{rownumber} & GPT-5~\cite{openai2025gpt5} & \modelapi{gpt-5} & \link{https://platform.openai.com/docs/models/gpt-5} \\
\stepcounter{rownumber}\arabic{rownumber} & GPT-5-mini~\cite{openai2025gpt5} & \modelapi{gpt-5-mini} & \link{https://platform.openai.com/docs/models/gpt-5-mini} \\

\stepcounter{rownumber}\arabic{rownumber} & GPT-OSS-120B~\cite{openai2025oss} & \company{open-source/}\modelapi{GPT-OSS-120B} & \link{https://huggingface.co/open-source/GPT-OSS-120B} \\
\stepcounter{rownumber}\arabic{rownumber} & Qwen3-235B-A22B-Instruct~\cite{qwen2025qwen3_235b_a22b} & \company{Qwen/}\modelapi{Qwen3-235B-A22B-Instruct} & \link{https://huggingface.co/Qwen/Qwen3-235B-A22B-Instruct} \\
\stepcounter{rownumber}\arabic{rownumber} & Llama-3.3-70B-Instruct-Turbo~\cite{meta2024llama3.3} & \company{meta-llama/}\modelapi{Llama-3.3-70B-Instruct-Turbo} & \link{https://huggingface.co/meta-llama/Llama-3.3-70B-Instruct} \\
\bottomrule
\end{tabular}
% \end{adjustbox}
\vspace{2mm}
\label{app:table_model_and_params}
\end{table}

\newpage
\section{Additional Experimental Results and Analysis}

\subsection{Detailed Metrics}
\label{app:exp1}

% ===== GPT-5 =====
\begin{table}[htbp]
\centering
\caption{Detailed metrics for GPT-5.}
\resizebox{0.75\textwidth}{!}{%
\begin{tabular}{l|ll|rrrrrrrrrrr}
\toprule
Size & Miss & Type & Insuf\% & Steps & Acc & TC & SuccTC & EffQ & EffRate & IG$_{\mu}$ & IR & IR$_{\text{eff}}$ & $\K$ \\
\midrule
\multirow{6}{*}{Small} & M1 & Normal & 0.3 & 2.5 & 99.0 & 1.5 & 1.5 & 1.3 & 0.94 & 0.670 & 1.51 & 1.33 & 1 \\
 &  & Only Fact & 0.7 & 2.5 & 98.7 & 1.5 & 1.5 & 1.3 & 0.95 & 0.676 & 1.49 & 1.33 & 1 \\
 &  & Only Relation & 0.0 & 2.5 & 100.0 & 1.5 & 1.5 & 1.4 & 0.98 & 0.681 & 1.46 & 1.37 & 1 \\
 & M2 & Normal & 0.5 & 4.6 & 100.0 & 3.6 & 3.6 & 2.7 & 0.85 & 0.601 & 1.79 & 1.33 & 2 \\
 &  & Only Fact & 0.5 & 4.6 & 100.0 & 3.6 & 3.6 & 2.7 & 0.87 & 0.603 & 1.78 & 1.36 & 2 \\
 &  & Only Relation & 0.5 & 4.2 & 100.0 & 3.2 & 3.1 & 2.7 & 0.93 & 0.642 & 1.59 & 1.37 & 2 \\
\midrule
\multirow{12}{*}{Medium} & M1 & Normal & 3.8 & 3.5 & 98.3 & 2.5 & 2.5 & 2.0 & 0.86 & 0.662 & 2.45 & 1.98 & 1 \\
 &  & Only Fact & 4.7 & 3.1 & 97.0 & 2.1 & 2.1 & 1.9 & 0.91 & 0.723 & 2.08 & 1.91 & 1 \\
 &  & Only Relation & 4.2 & 3.1 & 97.5 & 2.1 & 2.1 & 1.9 & 0.90 & 0.705 & 2.14 & 1.90 & 1 \\
 & M2 & Normal & 2.5 & 7.6 & 99.2 & 6.6 & 6.6 & 3.8 & 0.71 & 0.545 & 3.29 & 1.91 & 2 \\
 &  & Only Fact & 2.1 & 6.4 & 98.7 & 5.4 & 5.4 & 3.8 & 0.82 & 0.603 & 2.70 & 1.91 & 2 \\
 &  & Only Relation & 2.5 & 6.0 & 100.0 & 5.0 & 4.8 & 3.9 & 0.86 & 0.644 & 2.49 & 1.96 & 2 \\
 & M3 & Normal & 0.9 & 12.2 & 99.6 & 11.2 & 11.2 & 5.8 & 0.58 & 0.411 & 3.74 & 1.92 & 3 \\
 &  & Only Fact & 0.4 & 10.4 & 99.1 & 9.4 & 9.4 & 5.7 & 0.68 & 0.480 & 3.13 & 1.89 & 3 \\
 &  & Only Relation & 0.4 & 8.8 & 100.0 & 7.8 & 7.5 & 5.8 & 0.80 & 0.563 & 2.60 & 1.93 & 3 \\
 & M4 & Normal & 0.0 & 15.9 & 100.0 & 14.9 & 14.8 & 7.4 & 0.54 & 0.371 & 3.71 & 1.84 & 4 \\
 &  & Only Fact & 0.0 & 13.6 & 100.0 & 12.6 & 12.6 & 7.4 & 0.64 & 0.437 & 3.15 & 1.86 & 4 \\
 &  & Only Relation & 0.0 & 11.0 & 99.4 & 10.0 & 9.7 & 7.5 & 0.80 & 0.546 & 2.51 & 1.87 & 4 \\
\midrule
\multirow{6}{*}{Large} & M1 & Normal & 5.9 & 4.4 & 93.4 & 3.4 & 3.4 & 2.2 & 0.73 & 0.629 & 3.42 & 2.16 & 1 \\
 & M2 & Normal & 2.0 & 9.8 & 97.4 & 8.8 & 8.8 & 4.6 & 0.61 & 0.467 & 4.42 & 2.32 & 2 \\
 & M3 & Normal & 0.7 & 14.2 & 98.7 & 13.2 & 13.2 & 6.6 & 0.58 & 0.436 & 4.39 & 2.20 & 3 \\
 & M4 & Normal & 0.0 & 20.6 & 100.0 & 19.6 & 19.6 & 8.9 & 0.49 & 0.351 & 4.91 & 2.22 & 4 \\
 & M5 & Normal & 0.0 & 23.7 & 99.3 & 22.7 & 22.7 & 10.6 & 0.49 & 0.338 & 4.53 & 2.12 & 5 \\
 & M6 & Normal & 0.0 & 27.6 & 100.0 & 26.6 & 26.6 & 13.1 & 0.50 & 0.334 & 4.43 & 2.18 & 6 \\
 \bottomrule
\end{tabular}}
\label{tab:metrics_gpt_5}
\end{table}

% ===== GPT-5-mini =====
\begin{table}[htbp]
\centering
\caption{Detailed metrics for GPT-5-mini.}
\resizebox{0.75\textwidth}{!}{%
\begin{tabular}{l|ll|rrrrrrrrrrr}
\toprule
Size & Miss & Type & Insuf\% & Steps & Acc & TC & SuccTC & EffQ & EffRate & IG$_{\mu}$ & IR & IR$_{\text{eff}}$ & $\K$ \\
\midrule
\multirow{6}{*}{Small} & M1 & Normal & 0.0 & 2.4 & 100.0 & 1.4 & 1.4 & 1.3 & 0.99 & 0.711 & 1.36 & 1.34 & 1 \\
 &  & Only Fact & 0.3 & 2.3 & 99.7 & 1.3 & 1.3 & 1.3 & 0.99 & 0.714 & 1.34 & 1.33 & 1 \\
 &  & Only Relation & 0.3 & 2.5 & 99.7 & 1.5 & 1.4 & 1.4 & 0.95 & 0.668 & 1.49 & 1.37 & 1 \\
 & M2 & Normal & 2.6 & 4.2 & 99.5 & 2.8 & 2.8 & 2.7 & 0.97 & 0.686 & 1.39 & 1.33 & 2 \\
 &  & Only Fact & 1.1 & 3.8 & 100.0 & 2.8 & 2.8 & 2.7 & 0.99 & 0.696 & 1.39 & 1.36 & 2 \\
 &  & Only Relation & 1.1 & 3.9 & 98.4 & 2.9 & 2.8 & 2.7 & 0.94 & 0.658 & 1.45 & 1.33 & 2 \\
\midrule
\multirow{12}{*}{Medium} & M1 & Normal & 0.8 & 3.1 & 99.2 & 2.1 & 2.1 & 2.0 & 0.97 & 0.748 & 2.12 & 2.05 & 1 \\
 &  & Only Fact & 0.4 & 3.1 & 97.9 & 2.1 & 2.1 & 2.0 & 0.97 & 0.767 & 2.10 & 2.00 & 1 \\
 &  & Only Relation & 0.8 & 3.2 & 97.9 & 2.2 & 2.2 & 2.1 & 0.96 & 0.733 & 2.20 & 2.06 & 1 \\
 & M2 & Normal & 2.1 & 5.1 & 99.6 & 4.0 & 4.0 & 3.8 & 0.95 & 0.717 & 2.02 & 1.91 & 2 \\
 &  & Only Fact & 2.5 & 5.1 & 99.2 & 4.1 & 4.1 & 3.9 & 0.97 & 0.732 & 2.03 & 1.94 & 2 \\
 &  & Only Relation & 1.7 & 5.4 & 97.9 & 4.1 & 4.1 & 3.8 & 0.94 & 0.705 & 2.07 & 1.91 & 2 \\
 & M3 & Normal & 0.9 & 7.2 & 98.7 & 6.2 & 6.2 & 5.7 & 0.94 & 0.662 & 2.06 & 1.90 & 3 \\
 &  & Only Fact & 0.4 & 7.2 & 98.3 & 6.2 & 6.2 & 5.7 & 0.94 & 0.666 & 2.07 & 1.91 & 3 \\
 &  & Only Relation & 2.1 & 9.1 & 96.6 & 6.2 & 6.1 & 5.6 & 0.91 & 0.642 & 2.05 & 1.87 & 3 \\
 & M4 & Normal & 0.6 & 9.5 & 98.8 & 8.4 & 8.4 & 7.5 & 0.91 & 0.613 & 2.10 & 1.88 & 4 \\
 &  & Only Fact & 0.0 & 9.3 & 99.4 & 8.3 & 8.3 & 7.4 & 0.92 & 0.631 & 2.07 & 1.85 & 4 \\
 &  & Only Relation & 0.6 & 9.4 & 97.1 & 8.4 & 8.2 & 7.5 & 0.91 & 0.616 & 2.09 & 1.87 & 4 \\
\midrule
\multirow{6}{*}{Large} & M1 & Normal & 7.2 & 3.4 & 90.8 & 2.4 & 2.4 & 2.2 & 0.88 & 0.764 & 2.36 & 2.15 & 1 \\
 & M2 & Normal & 2.0 & 6.2 & 94.8 & 5.2 & 5.2 & 4.8 & 0.93 & 0.680 & 2.59 & 2.38 & 2 \\
 & M3 & Normal & 0.7 & 8.8 & 97.4 & 7.8 & 7.8 & 6.8 & 0.89 & 0.655 & 2.60 & 2.25 & 3 \\
 & M4 & Normal & 0.7 & 11.5 & 96.7 & 10.4 & 10.4 & 8.9 & 0.88 & 0.627 & 2.61 & 2.23 & 4 \\
 & M5 & Normal & 0.0 & 14.0 & 99.3 & 12.9 & 12.9 & 10.8 & 0.86 & 0.589 & 2.59 & 2.15 & 5 \\
 & M6 & Normal & 0.0 & 17.4 & 99.2 & 16.3 & 16.3 & 13.1 & 0.82 & 0.555 & 2.71 & 2.19 & 6 \\
\bottomrule
\end{tabular}}
\label{tab:metrics_gpt_5_mini}
\end{table}

% ===== Gemini-2.5-Flash =====
\begin{table}[htbp]
\centering
\caption{Detailed metrics for Gemini-2.5-Flash.}
\resizebox{0.88\textwidth}{!}{%
\begin{tabular}{l|ll|rrrrrrrrrrr}
\toprule
Size & Miss & Type & Insuf\% & Steps & Acc & TC & SuccTC & EffQ & EffRate & IG$_{\mu}$ & IR & IR$_{\text{eff}}$ & $\K$ \\
\midrule
\multirow{6}{*}{Small} & M1 & Normal & 9.3 & 3.2 & 92.1 & 2.2 & 2.1 & 1.2 & 0.68 & 0.480 & 2.18 & 1.20 & 1 \\
 &  & Only Fact & 7.3 & 4.4 & 93.0 & 3.0 & 2.1 & 1.2 & 0.53 & 0.386 & 2.96 & 1.23 & 1 \\
 &  & Only Relation & 9.6 & 2.9 & 90.1 & 1.8 & 1.6 & 1.2 & 0.75 & 0.527 & 1.77 & 1.19 & 1 \\
 & M2 & Normal & 9.5 & 6.1 & 89.5 & 5.1 & 4.9 & 2.4 & 0.63 & 0.434 & 2.54 & 1.21 & 2 \\
 &  & Only Fact & 6.8 & 8.1 & 91.1 & 6.2 & 4.9 & 2.5 & 0.51 & 0.360 & 3.08 & 1.27 & 2 \\
 &  & Only Relation & 8.9 & 5.4 & 85.8 & 4.2 & 3.8 & 2.4 & 0.69 & 0.470 & 2.12 & 1.22 & 2 \\
\midrule
\multirow{12}{*}{Medium} & M1 & Normal & 23.3 & 4.0 & 72.5 & 2.2 & 2.0 & 1.2 & 0.56 & 0.413 & 2.21 & 1.23 & 1 \\
 &  & Only Fact & 27.1 & 5.2 & 69.1 & 3.0 & 2.0 & 1.1 & 0.39 & 0.286 & 3.00 & 1.13 & 1 \\
 &  & Only Relation & 20.3 & 3.3 & 71.6 & 2.2 & 1.9 & 1.3 & 0.61 & 0.429 & 2.20 & 1.31 & 1 \\
 & M2 & Normal & 23.0 & 9.7 & 62.6 & 7.4 & 6.7 & 2.7 & 0.47 & 0.325 & 3.68 & 1.36 & 2 \\
 &  & Only Fact & 14.7 & 16.4 & 69.0 & 9.6 & 7.7 & 3.1 & 0.38 & 0.263 & 4.82 & 1.53 & 2 \\
 &  & Only Relation & 22.3 & 7.5 & 63.5 & 6.5 & 5.8 & 3.0 & 0.52 & 0.359 & 3.23 & 1.50 & 2 \\
 & M3 & Normal & 13.9 & 14.7 & 70.4 & 12.2 & 11.6 & 4.9 & 0.47 & 0.321 & 4.06 & 1.63 & 3 \\
 &  & Only Fact & 10.3 & 16.0 & 77.7 & 14.5 & 12.7 & 5.0 & 0.39 & 0.268 & 4.84 & 1.68 & 3 \\
 &  & Only Relation & 14.7 & 11.0 & 59.3 & 9.9 & 9.0 & 4.8 & 0.56 & 0.381 & 3.31 & 1.59 & 3 \\
 & M4 & Normal & 12.7 & 19.9 & 65.9 & 15.8 & 15.0 & 6.3 & 0.45 & 0.302 & 3.94 & 1.57 & 4 \\
 &  & Only Fact & 11.0 & 18.5 & 76.9 & 17.2 & 15.2 & 6.6 & 0.41 & 0.269 & 4.30 & 1.64 & 4 \\
 &  & Only Relation & 17.9 & 14.4 & 56.6 & 13.2 & 12.1 & 5.9 & 0.47 & 0.313 & 3.30 & 1.47 & 4 \\
\midrule
\multirow{6}{*}{Large} & M1 & Normal & 23.7 & 4.6 & 51.0 & 3.5 & 2.8 & 1.2 & 0.40 & 0.273 & 3.52 & 1.18 & 1 \\
 & M2 & Normal & 26.2 & 8.0 & 45.2 & 6.8 & 6.0 & 2.8 & 0.49 & 0.381 & 3.41 & 1.38 & 2 \\
 & M3 & Normal & 23.5 & 15.6 & 41.7 & 14.5 & 13.6 & 5.0 & 0.43 & 0.261 & 4.82 & 1.66 & 3 \\
 & M4 & Normal & 21.6 & 21.8 & 46.1 & 20.6 & 19.2 & 6.8 & 0.38 & 0.254 & 5.15 & 1.69 & 4 \\
 & M5 & Normal & 13.5 & 27.9 & 51.4 & 26.6 & 25.3 & 9.2 & 0.39 & 0.252 & 5.32 & 1.84 & 5 \\
 & M6 & Normal & 21.5 & 28.9 & 45.5 & 27.7 & 26.7 & 10.9 & 0.39 & 0.233 & 4.62 & 1.82 & 6 \\
\bottomrule
\end{tabular}}
\label{tab:metrics_gemini_2_5_flash}
\end{table}

% ===== Gemini-2.5-Pro =====
\begin{table}[htbp]
\centering
\caption{Detailed metrics for Gemini-2.5-Pro.}
\resizebox{0.88\textwidth}{!}{%
\begin{tabular}{l|ll|rrrrrrrrrrr}
\toprule
Size & Miss & Type & Insuf\% & Steps & Acc & TC & SuccTC & EffQ & EffRate & IG$_{\mu}$ & IR & IR$_{\text{eff}}$ & $\K$ \\
\midrule
\multirow{6}{*}{Small} & M1 & Normal & 1.0 & 2.4 & 99.0 & 1.4 & 1.4 & 1.3 & 0.96 & 0.691 & 1.39 & 1.31 & 1 \\
 &  & Only Fact & 2.0 & 2.4 & 98.3 & 1.4 & 1.4 & 1.3 & 0.96 & 0.684 & 1.36 & 1.28 & 1 \\
 &  & Only Relation & 1.3 & 2.3 & 97.4 & 1.3 & 1.3 & 1.3 & 0.97 & 0.685 & 1.34 & 1.29 & 1 \\
 & M2 & Normal & 3.7 & 4.0 & 91.6 & 2.9 & 2.9 & 2.5 & 0.90 & 0.617 & 1.47 & 1.27 & 2 \\
 &  & Only Fact & 4.2 & 3.9 & 92.6 & 2.9 & 2.9 & 2.5 & 0.92 & 0.641 & 1.45 & 1.27 & 2 \\
 &  & Only Relation & 5.8 & 3.9 & 88.9 & 2.8 & 2.8 & 2.5 & 0.92 & 0.632 & 1.42 & 1.27 & 2 \\
\midrule
\multirow{12}{*}{Medium} & M1 & Normal & 10.2 & 2.8 & 86.0 & 1.7 & 1.7 & 1.5 & 0.84 & 0.647 & 1.71 & 1.53 & 1 \\
 &  & Only Fact & 7.2 & 2.8 & 86.0 & 1.7 & 1.7 & 1.5 & 0.89 & 0.672 & 1.67 & 1.54 & 1 \\
 &  & Only Relation & 7.6 & 2.9 & 89.8 & 1.9 & 1.8 & 1.6 & 0.87 & 0.655 & 1.86 & 1.65 & 1 \\
 & M2 & Normal & 13.1 & 4.6 & 77.1 & 3.6 & 3.6 & 3.1 & 0.86 & 0.617 & 1.79 & 1.55 & 2 \\
 &  & Only Fact & 14.4 & 4.6 & 78.8 & 3.5 & 3.5 & 3.1 & 0.85 & 0.605 & 1.75 & 1.53 & 2 \\
 &  & Only Relation & 13.1 & 4.6 & 71.6 & 3.5 & 3.4 & 3.0 & 0.89 & 0.615 & 1.73 & 1.51 & 2 \\
 & M3 & Normal & 12.0 & 6.5 & 68.8 & 5.3 & 5.3 & 4.5 & 0.85 & 0.571 & 1.78 & 1.50 & 3 \\
 &  & Only Fact & 12.0 & 6.7 & 68.4 & 5.5 & 5.5 & 4.6 & 0.86 & 0.580 & 1.84 & 1.54 & 3 \\
 &  & Only Relation & 11.5 & 6.3 & 65.4 & 5.0 & 4.9 & 4.5 & 0.89 & 0.603 & 1.67 & 1.49 & 3 \\
 & M4 & Normal & 16.8 & 8.3 & 65.9 & 7.2 & 7.1 & 5.7 & 0.82 & 0.557 & 1.79 & 1.42 & 4 \\
 &  & Only Fact & 13.3 & 8.4 & 56.6 & 7.2 & 7.2 & 5.6 & 0.82 & 0.539 & 1.80 & 1.41 & 4 \\
 &  & Only Relation & 18.5 & 7.4 & 48.0 & 6.1 & 5.9 & 5.3 & 0.87 & 0.566 & 1.51 & 1.32 & 4 \\
\midrule
\multirow{6}{*}{Large} & M1 & Normal & 15.1 & 2.9 & 81.6 & 1.8 & 1.8 & 1.6 & 0.78 & 0.582 & 1.84 & 1.61 & 1 \\
 & M2 & Normal & 23.5 & 5.1 & 60.1 & 4.0 & 4.0 & 3.3 & 0.80 & 0.542 & 2.02 & 1.66 & 2 \\
 & M3 & Normal & 28.1 & 6.3 & 43.8 & 5.1 & 5.1 & 4.1 & 0.80 & 0.554 & 1.71 & 1.36 & 3 \\
 & M4 & Normal & 24.2 & 8.5 & 39.9 & 7.2 & 7.2 & 5.3 & 0.74 & 0.514 & 1.81 & 1.33 & 4 \\
 & M5 & Normal & 33.1 & 9.2 & 29.7 & 8.0 & 8.0 & 5.9 & 0.76 & 0.512 & 1.60 & 1.17 & 5 \\
 & M6 & Normal & 37.2 & 11.8 & 21.5 & 10.3 & 10.3 & 7.3 & 0.71 & 0.432 & 1.71 & 1.22 & 6 \\
\bottomrule
\end{tabular}}
\label{tab:metrics_gemini_2_5_pro}
\end{table}

% ===== Qwen3-235B =====
\begin{table}[htbp]
\centering
\caption{Detailed metrics for Qwen3-235B.}
\resizebox{0.88\textwidth}{!}{%
\begin{tabular}{l|ll|rrrrrrrrrrr}
\toprule
Size & Miss & Type & Insuf\% & Steps & Acc & TC & SuccTC & EffQ & EffRate & IG$_{\mu}$ & IR & IR$_{\text{eff}}$ & $\K$ \\
\midrule
\multirow{6}{*}{Small} & M1 & Normal & 2.6 & 5.0 & 96.4 & 1.7 & 1.7 & 1.3 & 0.87 & 0.607 & 1.70 & 1.29 & 1 \\
 &  & Only Fact & 1.3 & 4.9 & 97.4 & 1.6 & 1.6 & 1.3 & 0.91 & 0.656 & 1.56 & 1.28 & 1 \\
 &  & Only Relation & 3.7 & 5.9 & 95.9 & 1.8 & 1.5 & 1.3 & 0.81 & 0.564 & 1.84 & 1.25 & 1 \\
 & M2 & Normal & 1.6 & 8.4 & 95.8 & 4.0 & 3.9 & 2.6 & 0.79 & 0.548 & 1.99 & 1.32 & 2 \\
 &  & Only Fact & 1.6 & 7.6 & 97.9 & 3.3 & 3.3 & 2.6 & 0.89 & 0.616 & 1.64 & 1.32 & 2 \\
 &  & Only Relation & 2.6 & 8.5 & 92.6 & 3.8 & 3.1 & 2.5 & 0.80 & 0.552 & 1.88 & 1.25 & 2 \\
\midrule
\multirow{12}{*}{Medium} & M1 & Normal & 16.9 & 6.9 & 79.7 & 2.2 & 2.0 & 1.6 & 0.69 & 0.528 & 2.22 & 1.56 & 1 \\
 &  & Only Fact & 15.7 & 6.7 & 82.2 & 1.8 & 1.8 & 1.5 & 0.78 & 0.588 & 1.76 & 1.53 & 1 \\
 &  & Only Relation & 20.8 & 7.7 & 75.4 & 2.4 & 1.7 & 1.3 & 0.56 & 0.405 & 2.42 & 1.27 & 1 \\
 & M2 & Normal & 8.9 & 10.4 & 78.0 & 4.6 & 4.4 & 3.2 & 0.75 & 0.570 & 2.30 & 1.62 & 2 \\
 &  & Only Fact & 9.7 & 10.1 & 80.9 & 4.1 & 4.0 & 3.4 & 0.85 & 0.618 & 2.03 & 1.68 & 2 \\
 &  & Only Relation & 24.2 & 11.0 & 58.9 & 4.4 & 3.1 & 2.4 & 0.63 & 0.443 & 2.19 & 1.20 & 2 \\
 & M3 & Normal & 4.7 & 13.6 & 81.2 & 7.2 & 7.0 & 5.1 & 0.77 & 0.542 & 2.40 & 1.71 & 3 \\
 &  & Only Fact & 8.5 & 12.8 & 76.5 & 6.4 & 6.4 & 4.9 & 0.83 & 0.589 & 2.15 & 1.64 & 3 \\
 &  & Only Relation & 21.4 & 13.4 & 43.6 & 5.8 & 4.5 & 3.5 & 0.66 & 0.451 & 1.93 & 1.16 & 3 \\
 & M4 & Normal & 13.9 & 15.6 & 71.7 & 8.9 & 8.6 & 6.1 & 0.74 & 0.524 & 2.22 & 1.53 & 4 \\
 &  & Only Fact & 3.5 & 15.3 & 81.5 & 8.1 & 8.1 & 6.6 & 0.85 & 0.599 & 2.03 & 1.65 & 4 \\
 &  & Only Relation & 25.4 & 16.0 & 35.8 & 7.5 & 6.0 & 4.5 & 0.66 & 0.439 & 1.88 & 1.11 & 4 \\
\midrule
\multirow{6}{*}{Large} & M1 & Normal & 25.7 & 8.2 & 70.4 & 2.0 & 1.8 & 1.4 & 0.59 & 0.479 & 1.98 & 1.37 & 1 \\
 & M2 & Normal & 24.2 & 12.0 & 52.3 & 4.6 & 4.4 & 3.1 & 0.67 & 0.465 & 2.31 & 1.57 & 2 \\
 & M3 & Normal & 13.1 & 15.1 & 55.6 & 7.2 & 6.9 & 4.9 & 0.71 & 0.520 & 2.42 & 1.64 & 3 \\
 & M4 & Normal & 13.7 & 18.6 & 54.2 & 9.8 & 9.5 & 6.9 & 0.73 & 0.537 & 2.46 & 1.74 & 4 \\
 & M5 & Normal & 5.4 & 22.3 & 65.5 & 12.6 & 12.4 & 9.1 & 0.76 & 0.511 & 2.52 & 1.83 & 5 \\
 & M6 & Normal & 8.3 & 26.6 & 65.3 & 15.5 & 15.2 & 11.4 & 0.78 & 0.526 & 2.59 & 1.90 & 6 \\
\bottomrule
\end{tabular}}
\label{tab:metrics_Qwen3_235B}
\end{table}

% ===== Llama-3.3-70B =====
\begin{table}[htbp]
\centering
\caption{Detailed metrics for Llama-3.3-70B.}
\resizebox{0.88\textwidth}{!}{%
\begin{tabular}{l|ll|rrrrrrrrrrr}
\toprule
Size & Miss & Type & Insuf\% & Steps & Acc & TC & SuccTC & EffQ & EffRate & IG$_{\mu}$ & IR & IR$_{\text{eff}}$ & $\K$ \\
\midrule
\multirow{6}{*}{Small} & M1 & Normal & 6.6 & 4.4 & 66.6 & 3.3 & 3.3 & 1.1 & 0.34 & 0.239 & 3.33 & 1.08 & 1 \\
 &  & Only Fact & 6.0 & 4.1 & 66.2 & 3.1 & 3.1 & 1.1 & 0.40 & 0.279 & 3.09 & 1.14 & 1 \\
 &  & Only Relation & 11.6 & 3.9 & 61.3 & 2.8 & 2.7 & 0.9 & 0.35 & 0.244 & 2.84 & 0.94 & 1 \\
 & M2 & Normal & 5.3 & 5.9 & 53.7 & 4.9 & 4.8 & 2.1 & 0.44 & 0.314 & 2.43 & 1.06 & 2 \\
 &  & Only Fact & 3.7 & 5.6 & 54.2 & 4.5 & 4.5 & 2.2 & 0.51 & 0.357 & 2.24 & 1.08 & 2 \\
 &  & Only Relation & 5.8 & 5.0 & 46.3 & 4.0 & 3.9 & 1.7 & 0.43 & 0.296 & 2.01 & 0.87 & 2 \\
\midrule
\multirow{12}{*}{Medium} & M1 & Normal & 16.9 & 6.4 & 23.7 & 4.9 & 4.9 & 1.2 & 0.20 & 0.152 & 4.95 & 1.18 & 1 \\
 &  & Only Fact & 12.7 & 6.2 & 23.4 & 4.7 & 4.6 & 1.2 & 0.25 & 0.176 & 4.68 & 1.22 & 1 \\
 &  & Only Relation & 14.4 & 6.2 & 15.7 & 4.5 & 4.4 & 1.0 & 0.19 & 0.127 & 4.50 & 0.99 & 1 \\
 & M2 & Normal & 10.2 & 7.3 & 16.1 & 5.9 & 5.9 & 2.2 & 0.37 & 0.272 & 2.96 & 1.10 & 2 \\
 &  & Only Fact & 6.4 & 7.3 & 18.2 & 5.9 & 5.8 & 2.3 & 0.39 & 0.282 & 2.93 & 1.14 & 2 \\
 &  & Only Relation & 8.9 & 7.4 & 12.7 & 5.6 & 5.4 & 1.9 & 0.33 & 0.219 & 2.79 & 0.96 & 2 \\
 & M3 & Normal & 4.7 & 8.4 & 13.7 & 7.2 & 7.2 & 3.4 & 0.47 & 0.313 & 2.40 & 1.13 & 3 \\
 &  & Only Fact & 6.8 & 8.2 & 15.4 & 6.7 & 6.6 & 3.4 & 0.51 & 0.338 & 2.22 & 1.14 & 3 \\
 &  & Only Relation & 11.1 & 7.6 & 6.4 & 6.1 & 5.9 & 2.6 & 0.39 & 0.251 & 2.05 & 0.86 & 3 \\
 & M4 & Normal & 7.5 & 9.3 & 11.6 & 8.0 & 7.9 & 4.2 & 0.52 & 0.356 & 2.00 & 1.05 & 4 \\
 &  & Only Fact & 4.0 & 9.3 & 9.8 & 8.0 & 7.9 & 4.5 & 0.57 & 0.373 & 2.00 & 1.13 & 4 \\
 &  & Only Relation & 17.9 & 8.2 & 4.6 & 6.6 & 6.3 & 3.2 & 0.44 & 0.294 & 1.65 & 0.80 & 4 \\
\midrule
\multirow{6}{*}{Large} & M1 & Normal & 17.8 & 10.5 & 3.3 & 7.7 & 7.6 & 1.3 & 0.15 & 0.116 & 7.67 & 1.29 & 1 \\
 & M2 & Normal & 15.7 & 11.2 & 2.6 & 8.6 & 8.5 & 2.6 & 0.30 & 0.226 & 4.30 & 1.28 & 2 \\
 & M3 & Normal & 24.2 & 11.0 & 2.0 & 8.4 & 8.4 & 3.1 & 0.35 & 0.245 & 2.81 & 1.04 & 3 \\
 & M4 & Normal & 12.4 & 12.0 & 3.3 & 9.9 & 9.8 & 4.3 & 0.42 & 0.276 & 2.47 & 1.07 & 4 \\
 & M5 & Normal & 10.8 & 11.9 & 0.0 & 9.8 & 9.8 & 4.8 & 0.47 & 0.302 & 1.97 & 0.96 & 5 \\
 & M6 & Normal & 16.5 & 11.8 & 0.0 & 10.2 & 10.1 & 5.8 & 0.53 & 0.348 & 1.69 & 0.97 & 6 \\
\bottomrule
\end{tabular}}
\label{tab:metrics_Llama_3_3_70B}
\end{table}

% ===== GPT-OSS-120B =====
\begin{table}[htbp]
\centering
\caption{Detailed metrics for GPT-OSS-120B.}
\resizebox{0.88\textwidth}{!}{%
\begin{tabular}{l|ll|rrrrrrrrrrr}
\toprule
Size & Miss & Type & Insuf\% & Steps & Acc & TC & SuccTC & EffQ & EffRate & IG$_{\mu}$ & IR & IR$_{\text{eff}}$ & $\K$ \\
\midrule
\multirow{6}{*}{Small} & M1 & Normal & 46.0 & 3.2 & 71.5 & 1.0 & 0.9 & 0.7 & 0.45 & 0.328 & 1.04 & 0.73 & 1 \\
 &  & Only Fact & 44.7 & 2.4 & 72.8 & 1.4 & 1.0 & 0.7 & 0.39 & 0.278 & 1.36 & 0.74 & 1 \\
 &  & Only Relation & 34.4 & 2.1 & 80.8 & 1.0 & 1.0 & 0.9 & 0.61 & 0.440 & 1.01 & 0.88 & 1 \\
 & M2 & Normal & 44.2 & 5.4 & 56.8 & 2.2 & 1.9 & 1.6 & 0.50 & 0.346 & 1.12 & 0.79 & 2 \\
 &  & Only Fact & 48.9 & 5.6 & 55.8 & 2.4 & 1.8 & 1.4 & 0.37 & 0.256 & 1.22 & 0.69 & 2 \\
 &  & Only Relation & 47.4 & 2.8 & 61.1 & 1.8 & 1.8 & 1.5 & 0.55 & 0.383 & 0.89 & 0.76 & 2 \\
\midrule
\multirow{12}{*}{Medium} & M1 & Normal & 29.2 & 6.3 & 66.5 & 2.5 & 2.0 & 1.2 & 0.44 & 0.339 & 2.47 & 1.17 & 1 \\
 &  & Only Fact & 26.7 & 12.9 & 68.6 & 3.1 & 2.2 & 1.2 & 0.39 & 0.281 & 3.08 & 1.20 & 1 \\
 &  & Only Relation & 25.0 & 2.5 & 66.5 & 1.5 & 1.4 & 1.2 & 0.65 & 0.473 & 1.46 & 1.17 & 1 \\
 & M2 & Normal & 30.1 & 8.8 & 54.2 & 4.3 & 3.6 & 2.4 & 0.53 & 0.363 & 2.15 & 1.20 & 2 \\
 &  & Only Fact & 32.2 & 12.8 & 57.6 & 4.7 & 3.7 & 2.4 & 0.48 & 0.339 & 2.35 & 1.22 & 2 \\
 &  & Only Relation & 28.0 & 5.8 & 58.5 & 3.3 & 3.2 & 2.6 & 0.75 & 0.511 & 1.65 & 1.28 & 2 \\
 & M3 & Normal & 30.8 & 12.4 & 41.9 & 5.8 & 5.1 & 3.3 & 0.49 & 0.349 & 1.93 & 1.09 & 3 \\
 &  & Only Fact & 35.0 & 14.3 & 47.4 & 5.6 & 4.8 & 3.2 & 0.50 & 0.330 & 1.87 & 1.08 & 3 \\
 &  & Only Relation & 23.9 & 7.5 & 57.3 & 5.2 & 5.2 & 4.0 & 0.72 & 0.484 & 1.75 & 1.35 & 3 \\
 & M4 & Normal & 39.9 & 8.6 & 32.4 & 5.7 & 5.3 & 3.5 & 0.54 & 0.363 & 1.43 & 0.87 & 4 \\
 &  & Only Fact & 35.3 & 17.2 & 40.5 & 7.1 & 5.9 & 4.0 & 0.47 & 0.295 & 1.78 & 0.99 & 4 \\
 &  & Only Relation & 34.1 & 6.8 & 45.1 & 5.8 & 5.7 & 4.4 & 0.69 & 0.472 & 1.44 & 1.11 & 4 \\
\midrule
\multirow{6}{*}{Large} & M1 & Normal & 33.6 & 8.4 & 40.8 & 3.4 & 3.0 & 1.2 & 0.32 & 0.217 & 3.42 & 1.17 & 1 \\
 & M2 & Normal & 40.5 & 17.1 & 22.9 & 4.7 & 4.1 & 2.1 & 0.37 & 0.250 & 2.36 & 1.05 & 2 \\
 & M3 & Normal & 39.9 & 12.2 & 19.0 & 6.1 & 5.5 & 3.1 & 0.44 & 0.296 & 2.05 & 1.02 & 3 \\
 & M4 & Normal & 40.5 & 19.0 & 15.7 & 7.4 & 6.8 & 3.7 & 0.47 & 0.288 & 1.84 & 0.92 & 4 \\
 & M5 & Normal & 38.5 & 22.6 & 16.2 & 8.8 & 7.9 & 4.4 & 0.46 & 0.312 & 1.75 & 0.89 & 5 \\
 & M6 & Normal & 49.6 & 30.1 & 16.5 & 9.0 & 8.4 & 4.7 & 0.39 & 0.244 & 1.50 & 0.78 & 6 \\
\bottomrule
\end{tabular}}
\label{tab:metrics_gpt_oss_120b}
\end{table}

\newpage

\subsection{Calibration Against an Oracle Adaptive Policy}
\label{app:oracle_calibration}
Although our main metrics use $\K$ as a constructive reference budget, we also estimate the expected query complexity of an oracle adaptive policy as a calibration check. The oracle performs decision-tree search over legal tool calls and selects the query that minimizes the expected remaining number of queries:
\[
q^\star(\mathcal C_t)=
\arg\min_{q}
\Big[
1+
p_{\mathrm{true}}(q)\,V(\mathcal S_t^{q})
+
p_{\mathrm{false}}(q)\,V(\mathcal S_t^{\neg q})
\Big],
\]
where $\mathcal S_t^{q}$ and $\mathcal S_t^{\neg q}$ denote the candidate sets after observing a true/false answer to $q$, and $V(\cdot)$ is the expected number of remaining queries under the same policy. On instances with $\K=1,2,3$, the mean expected oracle query counts are $1.10$, $2.15$, and $3.18$, respectively. Thus, the oracle adaptive baseline we implement yields expected query counts close to the constructive reference budget, while current models remain far above both. We therefore use $\K$ as the primary metric denominator because it is instance-level, construction-guaranteed, and easy to interpret; we report the oracle numbers only as a calibration point, not as a substitute for $\K$.

\subsection{Insufficient Rate Across Different Environment Types}
Table~\ref{tab:insuf} reports the insufficient rate, i.e., cases where the number of valid tool calls falls below $\K$ and the corresponding accuracy gap between remaining and insufficient subsets.
Overall, insufficient rates increase with problem difficulty, rising from Small to Medium/Large and from lower to higher missing-clue levels, indicating that harder settings more often prevent agents from acquiring even a minimal amount of decisive information. Tool restrictions amplify this effect: across most models, only-relation tends to yield higher insufficient rates than normal or only-fact, especially for Medium and Large puzzles, suggesting that reliably extracting and exploiting relational constraints is a key bottleneck. In contrast, GPT-5 and GPT-5-mini maintain near-zero insufficient rates across evaluated settings, while weaker models (notably GPT-OSS-120B and Llama-3.3-70B) exhibit substantial insufficiency even at moderate difficulty. Notably, GPT-OSS-120B is an exception to the monotonic “harder $\rightarrow$ more insufficient” trend, showing very high insufficient rates even on Small puzzles, indicating an overconfident strategy that attempts to solve without adequately leveraging tools. Together, these results indicate that many failures stem not only from downstream reasoning, but from early breakdowns in information acquisition under uncertainty.
\begin{table*}[htbp]
\caption{For each model, we list three environment rows: Normal(N)/Only Fact(OF)/Only Relation(OR). Each missing-clue level (Mi) reports two columns: Insufficient Rate(\%) where valid tool calls are below $\K$ and the accuracy gap $\Delta$Acc between the remaining and insufficient subsets (Rem--Insuf). Large settings are evaluated under Normal only; entries for F/R are marked as --.}
\centering
\resizebox{\linewidth}{!}{%
\begin{tabular}{l c|cc|cc|cc|cc|cc|cc|cc|cc|cc|cc|cc|cc}
\toprule
\multirow{3}{*}{Model} & \multirow{3}{*}{Env}
& \multicolumn{4}{c|}{\textbf{Small}}
& \multicolumn{8}{c|}{\textbf{Medium}}
& \multicolumn{12}{c}{\textbf{Large}} \\
& 
& \multicolumn{2}{c}{M1} & \multicolumn{2}{c|}{M2}
& \multicolumn{2}{c}{M1} & \multicolumn{2}{c}{M2} & \multicolumn{2}{c}{M3} & \multicolumn{2}{c|}{M4}
& \multicolumn{2}{c}{M1} & \multicolumn{2}{c}{M2} & \multicolumn{2}{c}{M3} & \multicolumn{2}{c}{M4} & \multicolumn{2}{c}{M5} & \multicolumn{2}{c}{M6} \\
& 
& Insuf & $\Delta$Acc & Insuf & $\Delta$Acc
& Insuf & $\Delta$Acc & Insuf & $\Delta$Acc & Insuf & $\Delta$Acc & Insuf & $\Delta$Acc
& Insuf & $\Delta$Acc & Insuf & $\Delta$Acc & Insuf & $\Delta$Acc & Insuf & $\Delta$Acc & Insuf & $\Delta$Acc & Insuf & $\Delta$Acc \\
\midrule

% ========================= GPT-5 =========================
\multirow{3}{*}{GPT-5}
& N
& 0.3 & -1.0 & 0.5 & +0.0
& 3.8 & +32.9 & 2.5 & +33.3 & 0.9 & +50.0 & 0.0 & --
& 5.9 & +63.9 & 2.0 & +99.3 & 0.7 & +99.3 & 0.0 & -- & 0.0& -- & 0.0 & -- \\
& OF
& 0.7 & -1.3 & 0.5 & +0.0
& 4.7 & +25.5 & 2.1 & +39.6 & 0.4 & +99.6 & 0.0 & --
& -- & -- & -- & -- & -- & -- & -- & -- & -- & -- & -- & -- \\

& OR
& 0.0 & --   & 0.5 & +0.0
& 4.2 & +28.7 & 2.5 & +0.0  & 0.4 & +0.0  & 0.0 & --
& -- & -- & -- & -- & -- & -- & -- & -- & -- & -- & -- & -- \\
\midrule

% ========================= GPT-5-mini =========================
\multirow{3}{*}{GPT-5-mini}
& N
& 0.0 & --   & 2.6 & -0.5
& 0.8 & +49.6 & 2.1 & +20.0 & 0.9 & +49.1 & 0.6 & -1.2
& 7.2 & +58.7 & 2.0 & +62.7 & 0.7 & -2.6  & 0.7 & +97.4 & 0.0 & -- & 0.0 & -- \\
& OF
& 0.3 & +100.0 & 1.1 & +0.0
& 0.4 & -2.1   & 2.5 & -0.9  & 0.4 & -1.7  & 0.0 & --
& -- & -- & -- & -- & -- & -- & -- & -- & -- & -- & -- & -- \\
& OR
& 0.3 & -0.3 & 1.1 & +48.9
& 0.8 & +48.3 & 1.7 & -2.2  & 2.1 & +78.3 & 0.6 & -2.9
& -- & -- & -- & -- & -- & -- & -- & -- & -- & -- & -- & -- \\
\midrule

% ========================= Gemini-2.5-Flash =========================
\multirow{3}{*}{Gemini-2.5-Flash}
& N
& 9.3 & +46.4 & 9.5 & +55.9
& 23.3 & +35.2 & 23.0 & +52.4 & 13.9 & +67.3 & 12.7 & +65.1
& 23.7 & +31.7 & 26.2 & +50.0 & 23.5 & +48.3 & 21.6 & +58.3 & 13.5 & +53.6 & 21.5 & +57.9 \\
& OF
& 7.3 & +51.3 & 6.8 & +73.0
& 27.1 & +41.2 & 14.7 & +53.2 & 10.3 & +77.4 & 11.0 & +74.5
& -- & -- & -- & -- & -- & -- & -- & -- & -- & -- & -- & -- \\
& OR
& 9.6 & +53.9 & 8.9 & +55.5
& 20.3 & +48.0 & 22.3 & +47.1 & 14.7 & +55.7 & 17.9 & +69.0
& -- & -- & -- & -- & -- & -- & -- & -- & -- & -- & -- & -- \\
\midrule

% ========================= Gemini-2.5-Pro =========================
\multirow{3}{*}{Gemini-2.5-Pro}
& N
& 1.0 & +32.7 & 3.7 & +65.4
& 10.2 & +40.1 & 13.1 & +51.6 & 12.0 & +61.9 & 16.8 & +75.0
& 15.1 & +34.6 & 23.5 & +71.4 & 28.1 & +51.2 & 24.2 & +52.6 & 33.1 & +44.4 & 37.2 & +34.2 \\
& OF
& 2.0 & +49.3 & 4.2 & +70.6
& 7.2 & +35.6 & 14.4 & +57.7 & 12.0 & +65.5 & 13.3 & +55.3
& -- & -- & -- & -- & -- & -- & -- & -- & -- & -- & -- & -- \\
& OR
& 1.3 & +48.0 & 5.8 & +65.5
& 7.6 & +31.1 & 13.1 & +34.2 & 11.5 & +61.4 & 18.5 & +55.0
& -- & -- & -- & -- & -- & -- & -- & -- & -- & -- & -- & -- \\
\midrule

% ========================= Qwen3-235B =========================
\multirow{3}{*}{Qwen3-235B}
& N
& 2.6 & +60.5 & 1.6 & +97.3
& 16.9 & +47.8 & 8.9 & +59.4 & 4.7 & +66.1 & 13.9 & +73.5
& 25.7 & +36.1 & 24.2 & +51.1 & 13.1 & +52.4 & 13.7 & +57.4 & 5.4 & +69.3 & 8.3 & +71.2 \\
& OF
& 1.3 & +73.3 & 1.6 & +99.5
& 15.7 & +43.0 & 9.7 & +36.7 & 8.5 & +72.7 & 3.5 & +84.4
& -- & -- & -- & -- & -- & -- & -- & -- & -- & -- & -- & -- \\
& OR
& 3.7 & +61.9 & 2.6 & +33.5
& 20.8 & +38.5 & 24.2 & +43.0 & 21.4 & +32.5 & 25.4 & +38.9
& -- & -- & -- & -- & -- & -- & -- & -- & -- & -- & -- & -- \\
\midrule

% ========================= Llama-3.3-70B =========================
\multirow{3}{*}{Llama-3.3-70B}
& N
& 6.6 & +39.1 & 5.3 & +46.1
& 16.9 & +13.5 & 10.2 & +4.0 & 4.7 & +14.3 & 7.5 & +12.5
& 17.8 & +4.0 & 15.7 & +3.1 & 24.2 & +2.6 & 12.4 & +3.7 & 10.8 & +0.0 & 16.5 & +0.0 \\
& OF
& 6.0 & +29.1 & 3.7 & +11.8
& 12.7 & -0.0 & 6.4 & +5.2 & 6.8 & +3.1 & 4.0 & +10.2
& -- & -- & -- & -- & -- & -- & -- & -- & -- & -- & -- & -- \\
& OR
& 11.6 & +37.0 & 5.8 & +49.2
& 14.4 & -5.7 & 8.9 & +8.7 & 11.1 & +2.9 & 17.9 & +5.6
& -- & -- & -- & -- & -- & -- & -- & -- & -- & -- & -- & -- \\
\midrule

% ========================= GPT-OSS-120B =========================
\multirow{3}{*}{GPT-OSS-120B}
& N
& 46.0 & +45.9 & 44.2 & +78.4
& 29.2 & +38.7 & 30.1 & +59.4 & 30.8 & +54.5 & 39.9 & +49.0
& 33.6 & +26.0 & 40.5 & +24.9 & 39.9 & +28.8 & 40.5 & +26.4 & 38.5 & +23.5 & 49.6 & +32.8 \\
& OF
& 44.7 & +46.0 & 48.9 & +62.9
& 26.7 & +35.2 & 32.2 & +63.7 & 35.0 & +69.3 & 35.3 & +62.5
& -- & -- & -- & -- & -- & -- & -- & -- & -- & -- & -- & -- \\
& OR
& 34.4 & +42.6 & 47.4 & +78.0
& 25.0 & +39.0 & 28.0 & +51.7 & 23.9 & +65.9 & 34.1 & +68.4
& -- & -- & -- & -- & -- & -- & -- & -- & -- & -- & -- & -- \\
\bottomrule
\end{tabular}
}
\label{tab:insuf}
\end{table*}

\subsection{Budget Constraints}
\label{app:exp2}
As shown in Table~\ref{tab:exp2}, this section reports the budget setup and the corresponding results, comparing task performance and tool-use behavior across Tight/Normal/Relaxed regimes against the unconstrained Baseline.

\begin{table*}[htbp]
\centering
\caption{Medium puzzle performance across models, budgets, and missing clues ($\K$). 
For each model: Acc = accuracy (\%), Budget = tool call limit, Tool = avg tool calls, SuccTool = avg valid tool calls, EffTool = avg effective queries.}
\label{tab:exp2}
\resizebox{0.88\textwidth}{!}{%
\begin{tabular}{cc|cccc|cccc|cccc|cccc}
\toprule
&
& \multicolumn{4}{c|}{$\K=1$}
& \multicolumn{4}{c|}{$\K=2$}
& \multicolumn{4}{c|}{$\K=3$}
& \multicolumn{4}{c}{$\K=4$} \\
Model & Metric
& Baseline & Tight & Normal & Relaxed
& Baseline & Tight & Normal & Relaxed
& Baseline & Tight & Normal & Relaxed
& Baseline & Tight & Normal & Relaxed \\
\midrule
\multirow{5}{*}{Gemini-2.5-Flash}
& Acc
& 72.1 & 45.1 & 65.5 & 65.8
& 61.4 & 32.6 & 55.9 & 67.4
& 70.4 & 21.6 & 54.3 & 59.8
& 65.5 & 14.5 & 50.0 & 60.2 \\
& Budget
& $\infty$ & 1 & 2 & 3
& $\infty$ & 2 & 6 & 8
& $\infty$ & 3 & 11 & 14
& $\infty$ & 4 & 15 & 19 \\
& Tool
& 2.1 & 1.5 & 1.7 & 1.9
& 6.2 & 2.4 & 4.6 & 5.6
& 12.2 & 3.5 & 9.2 & 10.5
& 15.3 & 4.6 & 13.0 & 14.7 \\
& SuccTool
& 1.8 & 1.2 & 1.3 & 1.5
& 5.7 & 1.9 & 4.1 & 5.0
& 11.6 & 2.7 & 8.6 & 9.8
& 14.5 & 4.0 & 12.1 & 13.8 \\
& EffTool
& 1.2 & 0.9 & 1.1 & 1.2
& 2.6 & 1.6 & 2.7 & 3.0
& 4.9 & 2.4 & 4.4 & 4.6
& 6.2 & 3.4 & 5.9 & 6.2 \\
\midrule
\multirow{5}{*}{Gemini-2.5-pro}
& Acc
& 86.0 & 66.1 & 81.4 & 86.9
& 77.1 & 39.8 & 68.6 & 80.5
& 68.8 & 23.9 & 59.4 & 72.2
& 65.9 & 12.1 & 50.9 & 67.6 \\
& Budget
& $\infty$ & 1 & 2 & 3
& $\infty$ & 2 & 4 & 6
& $\infty$ & 3 & 6 & 9
& $\infty$ & 4 & 8 & 12 \\
& Tool
& 1.7 & 0.9 & 1.5 & 1.7
& 3.6 & 1.9 & 3.0 & 3.9
& 5.3 & 2.7 & 4.8 & 6.1
& 7.2 & 3.4 & 6.7 & 8.1 \\
& SuccTool
& 1.7 & 0.9 & 1.5 & 1.7
& 3.6 & 1.9 & 3.0 & 3.8
& 5.3 & 2.7 & 4.8 & 6.1
& 7.1 & 3.4 & 6.6 & 8.1 \\
& EffTool
& 1.5 & 0.9 & 1.4 & 1.6
& 3.1 & 1.8 & 2.7 & 3.3
& 4.5 & 2.6 & 4.2 & 4.8
& 5.7 & 3.0 & 5.6 & 6.2 \\
\midrule
\multirow{5}{*}{gpt-5-mini}
& Acc
& 99.2 & 94.5 & 97.0 & 99.6
& 99.6 & 94.9 & 97.5 & 100.0
& 98.7 & 93.2 & 98.3 & 99.1
& 98.8 & 94.2 & 98.3 & 100.0 \\
& Budget
& $\infty$ & 1 & 2 & 3
& $\infty$ & 2 & 4 & 6
& $\infty$ & 3 & 6 & 9
& $\infty$ & 4 & 9 & 14 \\
& Tool
& 2.1 & 1.8 & 1.9 & 2.1
& 4.0 & 3.8 & 3.9 & 4.0
& 6.2 & 5.7 & 5.9 & 6.1
& 8.4 & 7.4 & 7.8 & 8.0 \\
& SuccTool
& 2.1 & 1.8 & 1.9 & 2.1
& 4.0 & 3.8 & 3.9 & 4.0
& 6.2 & 5.7 & 5.9 & 6.1
& 8.4 & 7.4 & 7.8 & 8.0 \\
& EffTool
& 2.0 & 1.8 & 1.9 & 2.0
& 3.8 & 3.7 & 3.8 & 3.9
& 5.7 & 5.6 & 5.7 & 5.7
& 7.5 & 7.0 & 7.4 & 7.4 \\
\midrule
\multirow{5}{*}{gpt-5}
& Acc
& 98.3 & 97.0 & 94.9 & 97.9
& 99.1 & 96.6 & 98.7 & 100.0
& 99.6 & 97.0 & 99.1 & 98.3
& 100.0 & 97.7 & 100.0 & 99.4 \\
& Budget
& $\infty$ & 1 & 2 & 4
& $\infty$ & 2 & 6 & 10
& $\infty$ & 3 & 11 & 14
& $\infty$ & 4 & 15 & 19 \\
& Tool
& 2.4 & 2.0 & 1.9 & 2.0
& 6.5 & 3.9 & 4.0 & 4.1
& 11.2 & 5.8 & 6.2 & 6.5
& 14.9 & 7.5 & 8.5 & 8.5 \\
& SuccTool
& 2.4 & 2.0 & 1.9 & 2.0
& 6.5 & 3.9 & 4.0 & 4.1
& 11.2 & 5.8 & 6.2 & 6.5
& 14.8 & 7.5 & 8.5 & 8.5 \\
& EffTool
& 2.0 & 1.9 & 1.9 & 1.9
& 3.8 & 3.8 & 3.8 & 3.8
& 5.8 & 5.7 & 5.7 & 5.7
& 7.4 & 7.2 & 7.4 & 7.3 \\
\midrule
\multirow{5}{*}{Qwen3-235B}
& Acc
& 79.7 & 58.5 & 73.7 & 86.0
& 78.0 & 33.1 & 69.1 & 78.4
& 81.2 & 23.9 & 65.0 & 80.8
& 71.7 & 16.2 & 66.5 & 78.0 \\
& Budget
& $\infty$ & 1 & 2 & 4
& $\infty$ & 2 & 5 & 8
& $\infty$ & 3 & 7 & 11
& $\infty$ & 4 & 9 & 14 \\
& Tool
& 2.0 & 0.9 & 1.5 & 2.1
& 4.6 & 2.0 & 3.8 & 4.6
& 7.2 & 3.1 & 5.9 & 7.1
& 8.9 & 4.1 & 7.8 & 9.2 \\
& SuccTool
& 1.8 & 0.9 & 1.5 & 2.1
& 4.4 & 1.9 & 3.7 & 4.5
& 7.0 & 2.9 & 5.8 & 7.0
& 8.6 & 3.9 & 7.6 & 9.1 \\
& EffTool
& 1.5 & 0.8 & 1.3 & 1.6
& 3.2 & 1.7 & 2.9 & 3.3
& 5.1 & 2.5 & 4.4 & 5.1
& 6.1 & 3.1 & 6.0 & 6.6 \\
\midrule
\multirow{5}{*}{Llama-3.3-70B}
& Acc
& 23.9 & 12.7 & 23.3 & 33.2
& 16.1 & 5.1 & 15.7 & 24.2
& 13.7 & 6.8 & 11.5 & 22.6
& 11.6 & 1.7 & 8.1 & 22.5 \\
& Budget
& $\infty$ & 1 & 5 & 9
& $\infty$ & 2 & 6 & 10
& $\infty$ & 3 & 7 & 11
& $\infty$ & 4 & 8 & 12 \\
& Tool
& 4.5 & 1.5 & 4.3 & 7.1
& 5.9 & 2.1 & 5.2 & 7.9
& 7.2 & 2.7 & 6.1 & 9.2
& 8.0 & 3.7 & 7.1 & 10.0 \\
& SuccTool
& 4.5 & 1.5 & 4.2 & 6.9
& 5.9 & 2.0 & 5.1 & 7.8
& 7.2 & 2.6 & 6.0 & 9.0
& 7.9 & 3.6 & 7.0 & 9.8 \\
& EffTool
& 1.1 & 0.5 & 1.1 & 1.5
& 2.2 & 0.9 & 2.0 & 2.7
& 3.4 & 1.3 & 3.0 & 4.1
& 4.2 & 2.1 & 3.9 & 5.1 \\
\midrule
\multirow{5}{*}{GPT-OSS-120B}
& Acc
& 66.2 & 48.7 & 62.3 & 62.3
& 54.2 & 30.9 & 46.2 & 51.3
& 41.9 & 16.2 & 32.5 & 29.5
& 32.4 & 12.7 & 20.8 & 19.7 \\
& Budget
& $\infty$ & 1 & 3 & 5
& $\infty$ & 2 & 5 & 8
& $\infty$ & 3 & 6 & 9
& $\infty$ & 4 & 6 & 11 \\
& Tool
& 2.4 & 1.0 & 1.6 & 1.9
& 4.3 & 2.1 & 3.0 & 3.5
& 5.8 & 2.9 & 4.6 & 3.8
& 5.7 & 3.7 & 4.2 & 3.8 \\
& SuccTool
& 2.0 & 0.8 & 1.3 & 1.5
& 3.6 & 1.6 & 2.6 & 3.1
& 5.1 & 2.4 & 4.0 & 3.3
& 5.3 & 3.1 & 3.8 & 3.5 \\
& EffTool
& 1.2 & 0.7 & 1.1 & 1.2
& 2.4 & 1.5 & 2.1 & 2.3
& 3.3 & 2.0 & 3.0 & 2.4
& 3.5 & 2.6 & 2.9 & 2.5 \\
\bottomrule
\end{tabular}%
}
\end{table*}

\subsection{Pricing Signal}
\subsubsection{Setup}
\label{app:exp3_setup}
This experiment tests whether agents adapt their tool-selection strategies in response to explicit economic signals. We assign a virtual token cost to each query type (Fact vs.\ Relation) and instruct the agent to solve the puzzle while minimizing total cost without sacrificing correctness. Concretely, each tool call incurs a cost determined by its query type, and we vary the relative pricing across conditions to create incentives that favor one query type over the other.

We evaluate a set of pricing conditions that include a balanced baseline (equal costs), asymmetric regimes where Fact queries are cheaper or more expensive than Relation queries, joint shifts where both costs move together, extreme asymmetries (Fact-Very-Cheap / Fact-Very-Expensive), and a Tool-Free control where both query types have zero cost. To account for differences in token-scale across models, we use model-specific base prices while keeping the same ratios across conditions. Tables~\ref{tab:exp3_pricing_gemini25flash} and~\ref{tab:exp3_pricing_qwen3_235b} report the full pricing schedules for Gemini-2.5-Flash and Qwen3-235B, respectively.
\begin{table}[t]
\centering
\caption{Exp3 pricing conditions for \textbf{Gemini-2.5-Flash}.}
\small
\begin{tabular}{lccc}
\toprule
\textbf{Condition} & \textbf{Fact Price} & \textbf{Relation Price} & \textbf{Fact:Relation} \\
\midrule
Baseline            & 500  & 500  & 1:1 \\
Fact-Cheap          & 250  & 500  & 1:2 \\
Fact-Expensive      & 1000 & 500  & 2:1 \\
Relation-Cheap      & 500  & 250  & 2:1 \\
Relation-Expensive  & 500  & 1000 & 1:2 \\
Both-Cheap          & 250  & 250  & 1:1 \\
Both-Expensive      & 1000 & 1000 & 1:1 \\
Fact-Very-Cheap     & 100  & 2000 & 1:20 \\
Fact-Very-Expensive & 2000 & 100  & 20:1 \\
Tool-Free           & 0    & 0    & 1:1 \\
\bottomrule
\end{tabular}
\label{tab:exp3_pricing_gemini25flash}
\end{table}

\begin{table}[t]
\caption{Exp3 pricing conditions for \textbf{Qwen3-235B}.}
\centering
\small
\begin{tabular}{lccc}
\toprule
\textbf{Condition} & \textbf{Fact Price} & \textbf{Relation Price} & \textbf{Fact:Relation} \\
\midrule
Baseline            & 250  & 250  & 1:1 \\
Fact-Cheap          & 125  & 250  & 1:2 \\
Fact-Expensive      & 500  & 250  & 2:1 \\
Relation-Cheap      & 250  & 125  & 2:1 \\
Relation-Expensive  & 250  & 500  & 1:2 \\
Both-Cheap          & 125  & 125  & 1:1 \\
Both-Expensive      & 500  & 500  & 1:1 \\
Fact-Very-Cheap     & 50   & 1000 & 1:20 \\
Fact-Very-Expensive & 1000 & 50   & 20:1 \\
Tool-Free           & 0    & 0    & 1:1 \\
\bottomrule
\end{tabular}
\label{tab:exp3_pricing_qwen3_235b}
\end{table}

\subsubsection{results}

Table~\ref{tab:exp3_gemini2.5flash} and Table~\ref{tab:exp3_Qwen3_235B} summarize how agents respond to pricing interventions in terms of accuracy and query composition, showing in general that agents adjust the mix of Fact vs.\ Relation queries in the expected direction, but the degree of adaptation and its impact on accuracy vary across models and difficulty levels.
% ========================= Gemini-2.5-Flash =========================
\begin{table}[htbp]
\caption{Exp3 pricing results for Gemini-2.5-Flash in the normal environment (Medium puzzles).}
\centering
\small
\begin{tabular}{l|l|l|cccc}
\toprule
Missing Clues & Condition &Accuracy & Fact Tool Count & Relation Tool Count & Total Tool Count \\
\midrule
\multirow{10}{*}{M1} & baseline            & 73.0 & 1.69 & 1.24 & 2.93 \\
                    & fact cheap          & 65.5 & 1.84 & 0.97 & 2.81 \\
                    & fact expensive      & 68.4 & 1.09 & 1.58 & 2.67 \\
                    & relation cheap      & 65.7 & 0.72 & 1.43 & 2.15 \\
                    & relation expensive  & 73.1 & 1.73 & 1.25 & 2.98 \\
                    & both cheap          & 66.5 & 1.58 & 1.01 & 2.59 \\
                    & both expensive      & 68.4 & 1.72 & 1.01 & 2.73 \\
                    & fact very cheap     & 70.2 & 2.14 & 0.54 & 2.68 \\
                    & fact very expensive & 70.9 & 0.77 & 1.78 & 2.55 \\
                    & tool free           & 75.5 & 1.75 & 1.47 & 3.22 \\
\midrule
\multirow{10}{*}{M2} & baseline            & 62.9 & 3.85 & 3.04 & 6.89 \\
                    & fact cheap          & 65.0 & 4.19 & 3.58 & 7.77 \\
                    & fact expensive      & 56.4 & 2.18 & 3.76 & 5.94 \\
                    & relation cheap      & 63.4 & 2.50 & 4.17 & 6.67 \\
                    & relation expensive  & 67.4 & 4.96 & 3.45 & 8.42 \\
                    & both cheap          & 65.9 & 4.96 & 3.43 & 8.39 \\
                    & both expensive      & 64.5 & 4.79 & 3.04 & 7.82 \\
                    & fact very cheap     & 62.3 & 7.00 & 2.15 & 9.16 \\
                    & fact very expensive & 66.7 & 1.55 & 5.06 & 6.60 \\
                    & tool free           & 71.4 & 4.89 & 4.50 & 9.40 \\
\midrule
\multirow{10}{*}{M3} & baseline            & 66.5 & 6.30 & 5.83 & 12.14 \\
                    & fact cheap          & 62.2 & 7.44 & 4.38 & 11.82 \\
                    & fact expensive      & 59.6 & 4.41 & 6.54 & 10.96 \\
                    & relation cheap      & 56.9 & 3.82 & 6.03 & 9.85 \\
                    & relation expensive  & 59.1 & 7.69 & 4.54 & 12.23 \\
                    & both cheap          & 62.4 & 5.48 & 6.37 & 11.85 \\
                    & both expensive      & 59.6 & 4.97 & 5.27 & 10.24 \\
                    & fact very cheap     & 66.4 & 9.55 & 3.97 & 13.52 \\
                    & fact very expensive & 61.7 & 1.64 & 8.31 & 9.95 \\
                    & tool free           & 62.2 & 6.08 & 6.43 & 12.51 \\
\midrule
\multirow{10}{*}{M4} & baseline            & 56.9 & 7.93 & 5.22 & 13.15 \\
                    & fact cheap          & 61.3 & 9.06 & 6.32 & 15.37 \\
                    & fact expensive      & 53.0 & 3.96 & 8.99 & 12.95 \\
                    & relation cheap      & 52.7 & 3.87 & 8.84 & 12.71 \\
                    & relation expensive  & 56.7 & 9.63 & 6.16 & 15.80 \\
                    & both cheap          & 54.8 & 8.26 & 6.03 & 14.29 \\
                    & both expensive      & 53.5 & 6.99 & 6.09 & 13.08 \\
                    & fact very cheap     & 64.7 & 10.53 & 3.84 & 14.37 \\
                    & fact very expensive & 55.3 & 3.10 & 9.67 & 12.77 \\
                    & tool free           & 71.4 & 8.62 & 8.35 & 16.97 \\
\bottomrule
\end{tabular}
\label{tab:exp3_gemini2.5flash}
\end{table}

% ========================= Qwen3-235B =========================
\begin{table}[htbp]
\centering
\caption{Exp3 pricing results for Qwen3-235B in the normal environment (Medium puzzles).}
\small
\begin{tabular}{l|l|l|cccc}
\toprule
Missing Clues & Condition &Accuracy & Fact Tool Count & Relation Tool Count & Total Tool Count \\
\midrule
\multirow{10}{*}{M1} & baseline            & 83.8 & 1.60 & 0.39 & 1.99 \\
                    & fact cheap          & 85.5 & 1.72 & 0.26 & 1.98 \\
                    & fact expensive      & 84.3 & 1.25 & 0.86 & 2.12 \\
                    & relation cheap      & 84.3 & 1.23 & 0.79 & 2.02 \\
                    & relation expensive  & 82.2 & 1.59 & 0.27 & 1.86 \\
                    & both cheap          & 83.1 & 1.61 & 0.32 & 1.93 \\
                    & both expensive      & 83.9 & 1.56 & 0.39 & 1.95 \\
                    & fact very cheap     & 84.7 & 1.83 & 0.14 & 1.97 \\
                    & fact very expensive & 83.1 & 0.88 & 1.06 & 1.94 \\
                    & tool free           & 85.1 & 1.79 & 0.44 & 2.23 \\
\midrule
\multirow{10}{*}{M2} & baseline            & 76.7 & 3.48 & 0.73 & 4.21 \\
                    & fact cheap          & 74.6 & 3.47 & 0.52 & 3.99 \\
                    & fact expensive      & 71.6 & 2.50 & 1.19 & 3.69 \\
                    & relation cheap      & 73.3 & 2.69 & 1.30 & 3.99 \\
                    & relation expensive  & 75.4 & 3.60 & 0.58 & 4.18 \\
                    & both cheap          & 75.0 & 3.37 & 0.72 & 4.09 \\
                    & both expensive      & 70.3 & 3.28 & 0.75 & 4.03 \\
                    & fact very cheap     & 77.5 & 3.74 & 0.32 & 4.06 \\
                    & fact very expensive & 70.8 & 2.01 & 1.63 & 3.64 \\
                    & tool free           & 78.4 & 3.81 & 0.79 & 4.60 \\
\midrule
\multirow{10}{*}{M3} & baseline            & 70.5 & 5.49 & 0.87 & 6.36 \\
                    & fact cheap          & 74.4 & 5.67 & 0.66 & 6.33 \\
                    & fact expensive      & 67.5 & 3.93 & 1.82 & 5.75 \\
                    & relation cheap      & 67.5 & 4.02 & 1.76 & 5.78 \\
                    & relation expensive  & 67.9 & 5.79 & 0.64 & 6.43 \\
                    & both cheap          & 69.1 & 5.36 & 1.04 & 6.40 \\
                    & both expensive      & 75.2 & 5.55 & 0.90 & 6.45 \\
                    & fact very cheap     & 76.9 & 6.20 & 0.55 & 6.75 \\
                    & fact very expensive & 61.6 & 2.71 & 2.67 & 5.38 \\
                    & tool free           & 73.5 & 6.08 & 0.97 & 7.05 \\
\midrule
\multirow{10}{*}{M4} & baseline            & 65.3 & 7.44 & 1.12 & 8.56 \\
                    & fact cheap          & 70.5 & 7.57 & 0.60 & 8.17 \\
                    & fact expensive      & 57.8 & 5.03 & 2.02 & 7.05 \\
                    & relation cheap      & 69.9 & 6.06 & 2.03 & 8.09 \\
                    & relation expensive  & 68.8 & 7.42 & 0.74 & 8.16 \\
                    & both cheap          & 66.5 & 7.27 & 0.95 & 8.22 \\
                    & both expensive      & 68.8 & 6.78 & 1.05 & 7.83 \\
                    & fact very cheap     & 73.4 & 7.79 & 0.39 & 8.18 \\
                    & fact very expensive & 53.2 & 3.92 & 3.16 & 7.08 \\
                    & tool free           & 78.0 & 8.21 & 1.04 & 9.25 \\
\bottomrule
\end{tabular}
\label{tab:exp3_Qwen3_235B}
\end{table}

\newpage
\section{Prompts}
\subsection{Exp1: Basic Performance}
\begin{promptbox}[System Prompt for Exp1: Basic (Normal Environment)]
You are a reasoning agent solving a Zebra puzzle.

\textbf{Your capabilities}
\begin{itemize}
    \item You may reason step by step, but all reasoning text must be wrapped inside \texttt{<think>...</think>}.
    \item If you cannot deduce the correct answer, you may issue a query to the environment to obtain information.
    \item Each query must be pure JSON, strictly following the query protocol, and wrapped inside \texttt{<query>...</query>}.
    \item After receiving a response, integrate it into your reasoning and continue (still inside \texttt{<think>}).
    \item Never output both \texttt{<query>} and \texttt{<solution>} in the same message.
    \item If the puzzle is not uniquely solved, do \emph{not} output \texttt{<solution>}.
\end{itemize}

\textbf{Query protocol}

\emph{Common definitions}
\begin{itemize}
    \item Entity object: \texttt{\{ "attr": "<string>", "value": "<string>" \}}
    \item \texttt{HOUSES} = \texttt{\{houses\}}, \quad
          \texttt{ATTRS} = \texttt{\{attrs\}}, \quad
          \texttt{DOMAIN} = \texttt{\{domain\}}
    \item \texttt{attr} and \texttt{value} strings are normalized at runtime.
    \item \texttt{type} $\in$ \{\texttt{"fact"}, \texttt{"relation"}\}
\end{itemize}

\emph{Fact queries (\texttt{type:"fact"})}
\begin{lstlisting}
{
  "type": "fact",
  "rel": "found_at",
  "house": "<one of HOUSES>",
  "attr": "<string>",
  "value": "<string>"
}
\end{lstlisting}

\emph{Relation queries (\texttt{type:"relation"})}

Allowed relations:
\texttt{same\_house}, \texttt{not\_at}, \texttt{direct\_left}, \texttt{direct\_right}, \texttt{side\_by\_side}, \texttt{left\_of}, \texttt{right\_of}, \texttt{one\_between}, \texttt{two\_between}

\begin{lstlisting}
{
  "type": "relation",
  "rel": "<allowed relation>",
  "lhs": { "attr": "<string>", "value": "<string>" },
  "rhs": { "attr": "<string>", "value": "<string>" }
}
\end{lstlisting}

\textbf{Output rules}
\begin{itemize}
    \item Wrap all reasoning text in \texttt{<think>...</think>}.
    \item Wrap every query JSON in \texttt{<query>...</query>}.
    \item Wrap the final solution in \texttt{<solution>...</solution>}.
    \item Do not output any content outside these tags.
\end{itemize}

\textbf{Final requirement}
Once solved, output the complete solution JSON:

\begin{lstlisting}
<solution>
{
  "header": {header},
  "rows": [
    ["1", "<Name>", "<...>", "..."],
    ...
  ]
}
</solution>
\end{lstlisting}

The header must match the attribute list exactly, the number of rows must equal $N$, and all values must be drawn from \texttt{DOMAIN}.
\end{promptbox}

\begin{promptbox}[System Prompt for Exp1: Basic (Fact-only Environment))]
You are a reasoning agent solving a Zebra puzzle.

\textbf{Your capabilities}
\begin{itemize}
    \item You may reason step by step, but all reasoning text must be wrapped inside \texttt{<think>...</think>}.
    \item If you cannot deduce the correct answer, you may issue a query to the environment to obtain information.
    \item Each query must be pure JSON, strictly following the query protocol, and wrapped inside \texttt{<query>...</query>}.
    \item After receiving a response, integrate it into your reasoning and continue (still inside \texttt{<think>}).
    \item Never output both \texttt{<query>} and \texttt{<solution>} in the same message.
    \item If the puzzle is not uniquely solved, do \emph{not} output \texttt{<solution>}.
\end{itemize}

\textbf{Query protocol}

\emph{Common definitions}
\begin{itemize}
    \item Entity object: \texttt{\{ "attr": "<string>", "value": "<string>" \}}
    \item \texttt{HOUSES} = \texttt{\{houses\}}, \quad
          \texttt{ATTRS} = \texttt{\{attrs\}}, \quad
          \texttt{DOMAIN} = \texttt{\{domain\}}
    \item \texttt{attr} and \texttt{value} strings are normalized at runtime.
    \item \texttt{type} $\in$ \{\texttt{"fact"}\}
\end{itemize}

\emph{Fact queries (\texttt{type:"fact"})}
\begin{lstlisting}
{
  "type": "fact",
  "rel": "found_at",
  "house": "<one of HOUSES>",
  "attr": "<string>",
  "value": "<string>"
}
\end{lstlisting}

\textbf{Output rules}
\begin{itemize}
    \item Wrap all reasoning text in \texttt{<think>...</think>}.
    \item Wrap every query JSON in \texttt{<query>...</query>}.
    \item Wrap the final solution in \texttt{<solution>...</solution>}.
    \item Do not output any content outside these tags.
\end{itemize}

\textbf{Final requirement}
Once solved, output the complete solution JSON:

\begin{lstlisting}
<solution>
{
  "header": {header},
  "rows": [
    ["1", "<Name>", "<...>", "..."],
    ...
  ]
}
</solution>
\end{lstlisting}

The header must match the attribute list exactly, the number of rows must equal $N$, and all values must be drawn from \texttt{DOMAIN}.
\end{promptbox}
\begin{promptbox}[System Prompt for Exp1: Basic (Relation-Only Environment)]
You are a reasoning agent solving a Zebra puzzle.

\textbf{Your capabilities}
\begin{itemize}
    \item You may reason step by step, but all reasoning text must be wrapped inside \texttt{<think>...</think>}.
    \item If you cannot deduce the correct answer, you may issue a query to the environment to obtain information.
    \item Each query must be pure JSON, strictly following the query protocol, and wrapped inside \texttt{<query>...</query>}.
    \item After receiving a response, integrate it into your reasoning and continue (still inside \texttt{<think>}).
    \item Never output both \texttt{<query>} and \texttt{<solution>} in the same message.
    \item If the puzzle is not uniquely solved, do \emph{not} output \texttt{<solution>}.
\end{itemize}

\textbf{Query protocol}

\emph{Common definitions}
\begin{itemize}
    \item Entity object: \texttt{\{ "attr": "<string>", "value": "<string>" \}}
    \item \texttt{HOUSES} = \texttt{\{houses\}}, \quad
          \texttt{ATTRS} = \texttt{\{attrs\}}, \quad
          \texttt{DOMAIN} = \texttt{\{domain\}}
    \item \texttt{attr} and \texttt{value} strings are normalized at runtime.
    \item \texttt{type} $\in$ \{ \texttt{"relation"}\}
\end{itemize}

\emph{Relation queries (\texttt{type:"relation"})}

Allowed relations:
\texttt{same\_house}, \texttt{not\_at}, \texttt{direct\_left}, \texttt{direct\_right}, \texttt{side\_by\_side}, \texttt{left\_of}, \texttt{right\_of}, \texttt{one\_between}, \texttt{two\_between}

\begin{lstlisting}
{
  "type": "relation",
  "rel": "<allowed relation>",
  "lhs": { "attr": "<string>", "value": "<string>" },
  "rhs": { "attr": "<string>", "value": "<string>" }
}
\end{lstlisting}

\textbf{Output rules}
\begin{itemize}
    \item Wrap all reasoning text in \texttt{<think>...</think>}.
    \item Wrap every query JSON in \texttt{<query>...</query>}.
    \item Wrap the final solution in \texttt{<solution>...</solution>}.
    \item Do not output any content outside these tags.
\end{itemize}

\textbf{Final requirement}
Once solved, output the complete solution JSON:

\begin{lstlisting}
<solution>
{
  "header": {header},
  "rows": [
    ["1", "<Name>", "<...>", "..."],
    ...
  ]
}
</solution>
\end{lstlisting}

The header must match the attribute list exactly, the number of rows must equal $N$, and all values must be drawn from \texttt{DOMAIN}.
\end{promptbox}

\subsection{Exp2: Budget Constraints}

\begin{promptbox}[System Prompt for Exp2: Budget Constraints]
You are a reasoning agent solving a Zebra puzzle.

\textbf{Your capabilities}
\begin{itemize}
    \item You may reason step by step, but all reasoning text must be wrapped inside \texttt{<think>...</think>}.
    \item If you cannot deduce the correct answer, you may issue a query to the environment to obtain information.
    \item Each query must be pure JSON, strictly following the query protocol, and wrapped inside \texttt{<query>...</query>}.
    \item After receiving a response, integrate it into your reasoning and continue (still inside \texttt{<think>}).
    \item Never output both \texttt{<query>} and \texttt{<solution>} in the same message.
    \item If the puzzle is not uniquely solved, do \emph{not} output \texttt{<solution>}.
\end{itemize}

\textbf{Query protocol}

\emph{Common definitions}
\begin{itemize}
    \item Entity object: \texttt{\{ "attr": "<string>", "value": "<string>" \}}
    \item \texttt{HOUSES} = \texttt{\{houses\}}, \quad
          \texttt{ATTRS} = \texttt{\{attrs\}}, \quad
          \texttt{DOMAIN} = \texttt{\{domain\}}
    \item \texttt{attr} and \texttt{value} strings are normalized at runtime.
    \item \texttt{type} $\in$ \{\texttt{"fact"}, \texttt{"relation"}\}
\end{itemize}

\emph{Fact queries (\texttt{type:"fact"})}
\begin{lstlisting}
{
  "type": "fact",
  "rel": "found_at",
  "house": "<one of HOUSES>",
  "attr": "<string>",
  "value": "<string>"
}
\end{lstlisting}

\emph{Relation queries (\texttt{type:"relation"})}

Allowed relations:
\texttt{same\_house}, \texttt{not\_at}, \texttt{direct\_left}, \texttt{direct\_right}, \texttt{side\_by\_side}, \texttt{left\_of}, \texttt{right\_of}, \texttt{one\_between}, \texttt{two\_between}

\begin{lstlisting}
{
  "type": "relation",
  "rel": "<allowed relation>",
  "lhs": { "attr": "<string>", "value": "<string>" },
  "rhs": { "attr": "<string>", "value": "<string>" }
}
\end{lstlisting}

\textbf{Output rules}
\begin{itemize}
    \item Wrap all reasoning text in \texttt{<think>...</think>}.
    \item Wrap every query JSON in \texttt{<query>...</query>}.
    \item Wrap the final solution in \texttt{<solution>...</solution>}.
    \item Do not output any content outside these tags.
\end{itemize}

\textbf{Final requirement}
Once solved, output the complete solution JSON:

\begin{lstlisting}
<solution>
{
  "header": {header},
  "rows": [
    ["1", "<Name>", "<...>", "..."],
    ...
  ]
}
</solution>
\end{lstlisting}

The header must match the attribute list exactly, the number of rows must equal $N$, and all values must be drawn from \texttt{DOMAIN}.

\textbf{Budget Constraint}

You have a budget of \textbf{Y} tool calls for this puzzle.
\begin{itemize}
 \item Each query (fact or relation) costs 1 from your budget.
 \item Invalid queries that fail validation also cost 1.
 \item This is a target - plan your queries carefully.
 \item  Current remaining budget will be shown after each query response.
\end{itemize}

\textbf{Strategy tips:}
\begin{itemize}
    \item Start by reasoning about what information would be most valuable.
    \item  Avoid redundant queries - track what you've already learned.
    \item  Prioritize queries that can eliminate the most possibilities.
    \item  If running low on budget, make your best guess based on available information.
\end{itemize}

\end{promptbox}

\begin{promptbox}[Budget Update Prompt]
[Budget: X/Y remaining]
\end{promptbox}

\subsection{Exp3: Pricing Signal}

\begin{promptbox}[System Prompt for Exp2: Budget Constraints]
You are a reasoning agent solving a Zebra puzzle.

\textbf{Your capabilities}
\begin{itemize}
    \item You may reason step by step, but all reasoning text must be wrapped inside \texttt{<think>...</think>}.
    \item If you cannot deduce the correct answer, you may issue a query to the environment to obtain information.
    \item Each query must be pure JSON, strictly following the query protocol, and wrapped inside \texttt{<query>...</query>}.
    \item After receiving a response, integrate it into your reasoning and continue (still inside \texttt{<think>}).
    \item Never output both \texttt{<query>} and \texttt{<solution>} in the same message.
    \item If the puzzle is not uniquely solved, do \emph{not} output \texttt{<solution>}.
\end{itemize}

\textbf{Query protocol}

\emph{Common definitions}
\begin{itemize}
    \item Entity object: \texttt{\{ "attr": "<string>", "value": "<string>" \}}
    \item \texttt{HOUSES} = \texttt{\{houses\}}, \quad
          \texttt{ATTRS} = \texttt{\{attrs\}}, \quad
          \texttt{DOMAIN} = \texttt{\{domain\}}
    \item \texttt{attr} and \texttt{value} strings are normalized at runtime.
    \item \texttt{type} $\in$ \{\texttt{"fact"}, \texttt{"relation"}\}
\end{itemize}

\emph{Fact queries (\texttt{type:"fact"})}
\begin{lstlisting}
{
  "type": "fact",
  "rel": "found_at",
  "house": "<one of HOUSES>",
  "attr": "<string>",
  "value": "<string>"
}
\end{lstlisting}

\emph{Relation queries (\texttt{type:"relation"})}

Allowed relations:
\texttt{same\_house}, \texttt{not\_at}, \texttt{direct\_left}, \texttt{direct\_right}, \texttt{side\_by\_side}, \texttt{left\_of}, \texttt{right\_of}, \texttt{one\_between}, \texttt{two\_between}

\begin{lstlisting}
{
  "type": "relation",
  "rel": "<allowed relation>",
  "lhs": { "attr": "<string>", "value": "<string>" },
  "rhs": { "attr": "<string>", "value": "<string>" }
}
\end{lstlisting}

\textbf{Output rules}
\begin{itemize}
    \item Wrap all reasoning text in \texttt{<think>...</think>}.
    \item Wrap every query JSON in \texttt{<query>...</query>}.
    \item Wrap the final solution in \texttt{<solution>...</solution>}.
    \item Do not output any content outside these tags.
\end{itemize}

\textbf{Final requirement}
Once solved, output the complete solution JSON:

\begin{lstlisting}
<solution>
{
  "header": {header},
  "rows": [
    ["1", "<Name>", "<...>", "..."],
    ...
  ]
}
</solution>
\end{lstlisting}

The header must match the attribute list exactly, the number of rows must equal $N$, and all values must be drawn from \texttt{DOMAIN}.

\textbf{ Tool Pricing \& Optimization Goal}
Each tool call has a token cost:
\begin{itemize}
    \item Fact query: {fact\_price} tokens
    \item Relation query: {relation\_price} tokens
\end{itemize}

\textbf{Your goal is to minimize total token cost while solving the puzzle accurately and reliably.}

\textbf{After each response, you will see your cumulative cost:
[Token usage: XXXX reasoning + YYYY tools = ZZZZ total]}

\textbf{Strategy guidelines:}
\begin{itemize}
\item Be efficient: Only query when necessary for deduction.
    \item Be strategic: Choose the query type that gives maximum information per token.
 \item Be accurate: Ensure your solution is correct - an incorrect solution wastes all spent tokens.
 \item Be reliable: When uncertain, query rather than guess incorrectly.

\end{itemize}
Balance efficiency with accuracy - the optimal strategy minimizes tokens while maintaining correctness.
\end{promptbox}

% \section{Technical appendices and supplementary material}
% Technical appendices with additional results, figures, graphs, and proofs may be submitted with the paper submission before the full submission deadline (see above). You can upload a ZIP file for videos or code, but do not upload a separate PDF file for the appendix. There is no page limit for the technical appendices. 

% Note: Think of the appendix as ``optional reading'' for reviewers. The paper must be able to stand alone without the appendix; for example, adding critical experiments that support the main claims to an appendix is inappropriate. 

%%%%%%%%%%%%%%%%%%%%%%%%%%%%%%%%%%%%%%%%%%%%%%%%%%%%%%%%%%%%
\end{document}